\pdfoutput=1
\documentclass{article}

\usepackage[final]{neurips_2020}

\usepackage[utf8]{inputenc} %
\usepackage[T1]{fontenc}    %
\usepackage{hyperref}       %
\usepackage{url}            %
\usepackage{booktabs}       %
\usepackage{amsfonts}       %
\usepackage{nicefrac}       %
\usepackage{microtype}      %
\usepackage{changes}

\usepackage{makecell}
\usepackage{paralist}
\usepackage{multirow}
\usepackage{subcaption}
\usepackage{xspace}
\usepackage{enumitem}
\usepackage{xcolor}
\usepackage{amsmath}

\newcommand{\blockcomment}[1]{}
\newcommand{\etal}[0]{\xspace\emph{et al.}\xspace}

\newcommand{\mL}[1]{\ensuremath\mathcal{L}_\mathrm{#1}}
\newcommand{\mE}[1]{\ensuremath\mathcal{E}_\mathrm{#1}}
\newcommand{\cossim}[2]{\ensuremath\frac{#1\cdot#2}{\|#1\|\|#2\|}}

\DeclareMathSizes{10}{9}{6.3}{4.5}
\makeatletter
\g@addto@macro\normalsize{%
\addtolength{\abovedisplayskip}{-5pt}
\addtolength{\belowdisplayskip}{-5pt}
\addtolength{\abovedisplayshortskip}{-9pt}
\addtolength{\belowdisplayshortskip}{-5pt}
}
\makeatother
\usepackage[font=small]{caption}
\graphicspath{{figures/}}

\title{Self-Learning Transformations for Improving Gaze and Head Redirection}

\author{Yufeng Zheng$^1$, 
\hspace{2mm}Seonwook Park$^1$, 
\hspace{2mm}Xucong Zhang$^1$, 
\hspace{2mm}Shalini De Mello$^2$, 
\hspace{2mm}Otmar Hilliges$^1$\\
$^1$Department of Computer Science, ETH Zurich 
\hspace{15mm} $^2$NVIDIA\\
\texttt{\{firstname.lastname\}@inf.ethz.ch} \hspace{2mm} \texttt{shalinig@nvidia.com}
}

\begin{document}

\maketitle

\begin{abstract}
Many computer vision tasks rely on labeled data. Rapid progress in generative modeling has led to the ability to synthesize photorealistic images. However, controlling specific aspects of the generation process such that the data can be used for supervision of downstream tasks remains challenging. In this paper we propose a novel generative model for images of faces, that is capable of producing high-quality images under fine-grained control over eye gaze and head orientation angles. This requires the disentangling of many appearance related factors including gaze and head orientation but also lighting, hue etc. 
We propose a novel architecture which learns to discover, disentangle and encode these extraneous variations in a self-learned manner. We further show that explicitly disentangling task-irrelevant factors results in more accurate modelling of gaze and head orientation.
A novel evaluation scheme shows that our method improves upon the state-of-the-art in redirection accuracy and disentanglement between gaze direction and head orientation changes. Furthermore, we show that in the presence of limited amounts of real-world training data, our method allows for improvements in the downstream task of semi-supervised cross-dataset gaze estimation. Please check our project page at: \href{https://ait.ethz.ch/projects/2020/STED-gaze/}{https://ait.ethz.ch/projects/2020/STED-gaze/} 
\end{abstract}

\section{Introduction}

Extracting information from images of human faces is one of the core problems in artificial intelligence and computer vision. 
For example, estimating eye gaze has many applications in the social sciences \cite{Oertel2015ICMI,Fan2019ICCV}, cognitive sciences \cite{buswell1935people,rothkopf2007task}, and can enable novel applications in graphics, HCI, AR and VR  \cite{Fridman2018CHI,Huang2016MM,Feit2017CHI,Smith2013UIST,Zhang2017UIST,Patney2016SIGGRAPH}.
Given the need for large amounts of training data for learning-based gaze estimation approaches, much attention has been given to synthesizing training data via a graphics pipeline \cite{Wood2015ICCV,Wood2016ETRA} and synthetic-to-real domain adaptation \cite{Shrivastava2017CVPR}. However, domain adaptation approaches can be sensitive to changes in the underlying distribution of gaze directions, producing unfaithful images that do not help in improving gaze estimator performance \cite{Lee2018ICLR}.
More recently, approaches to re-direct the gaze in real-images has emerged as an alternative approach to attain gaze estimation training data \cite{Ganin2016ECCV,Yu2019CVPR,He2019ICCV,Wood2016ETRA}. This task involves the learning of a mapping between two images with differing gaze directions and requires either paired synthetic \cite{Yu2019CVPR} or high-quality real-world images captured under controlled conditions \cite{He2019ICCV}.

However, leveraging gaze data from in-the-wild settings \cite{Krafka2016CVPR,Zhang2015CVPR} for this purpose has so far been elusive \cite{Ganin2016ECCV, Park2019ICCV}. The underlying factors that we wish to explicitly control (gaze, head orientation) are entangled with many other extraneous factors (e.g., lighting, hue, etc) that are typically not known \emph{a priori} (see Fig. \ref{fig:qualitative_main}, first and last columns). 
Conditional unpaired image-to-image translation methods such as StarGAN \cite{starGAN} and similar \cite{he2019attgan,pumarola2018ganimation,wu2019relgan} provide a promising framework to tackle this in-the-wild problem, where perfectly paired images are not available. However, accurately controlling the explicit factors to a degree where the generated data is useful for downstream tasks remains challenging.

To solve this challenge we propose a new approach for gaze and head orientation redirection along with a principled way to evaluate such methods, and 
demonstrate its effectiveness at improving the challenging downstream tasks of semi-supervised cross-dataset gaze estimation. 
Specifically, we design a novel framework that learns to simultaneously encode \emph{both} the explicit task-relevant (gaze, head orientation) and the extraneous, unlabeled task-irrelevant factors.
To do this, we propose a self-transforming encoder-decoder architecture with multiple transformable factors at the bottleneck. Each factor consists of a latent embedding and a self-predicted pseudo condition, which describes the amount of its variation as present in a particular image.
In addition, we propose several novel constraints to effectively disentangle the various independent factors, while maintaining precise control over the explicitly manipulated factors in the redirected images.

We introduce several evaluation schemes to assess the quality of redirection in a principled manner.
First, we propose a \emph{redirection error} metric which measures how accurately the target gaze direction and head orientation values are reproduced in the generated images.
Second, we propose a \emph{task disentanglement error} metric which measures how much the perceived gaze direction or head orientation changes when other factors are adjusted.
With these two metrics, we show that our novel architecture is effective in isolating a multitude of independent factors, and thus performs well in terms of faithfulness of redirection.
To further evaluate the impact of our proposed approaches, we demonstrate the effectiveness of our gaze redirection scheme on the
 task of semi-supervised cross-dataset gaze estimation. 
Here, we first train our gaze redirection network in a semi-supervised manner with a small amount of labeled training data. Then we augment the labeled data via redirection and train a gaze network on the augmented data to demonstrate improvements compared to training on only the non-augmented labeled data. %

In summary, we contribute:
\vspace{-1mm}
\begin{itemize}[leftmargin=*]
\item a novel self-transforming encoder-decoder architecture that learns to control both explicit and extraneous factors in a disentangled manner,
\vspace{-0.5mm}
\item a principled evaluation scheme for measuring the accuracy of gaze redirection methods, and the disentanglement between task-specific (explicit) and task-irrelevant (extraneous) factors,
\vspace{-0.5mm}
\item high-fidelity gaze and head orientation manipulation on the generated face images, and
\vspace{-0.5mm}
\item demonstration of performance improvements on four datasets in the real-world downstream tasks of cross-dataset gaze estimation, by augmenting real training data via redirection.
\end{itemize}

\section{Background}
\vspace{-2mm}%
\textbf{Gaze Redirection.}
DeepWarp is an early work that learns a warping map between pairs of eye images with different gaze directions \cite{Ganin2016ECCV}.
It only works for a limited range of target gaze directions.
Yu\etal extend it with synthetic data from \cite{wood2016learning} along with a gaze estimator \cite{Yu2019CVPR}. %
Warping-based methods cannot generate new image content since every pixel in the generated image is interpolated from the original input image.
Therefore, such methods cannot synthesize the change of lighting conditions or extreme gaze directions and head orientations. He\etal present a GAN-based architecture where perceptual loss \cite{johnson2016perceptual} and cycle consistency \cite{cycleGAN} are used to supervise the redirection process. However, their work uses high-resolution images with controlled lighting conditions and cannot generalize well to in-the-wild images \cite{He2019ICCV}. Alternatively, Wood\etal \cite{wood2018gazedirector} fit morphable models to eye regions to redirect them, but are limited by the fidelity and flexibility of the morphable model. FAZE uses a disentangling encoder-decoder architecture to learn a latent representation to rotate gaze and head orientation, but its redirection yields images of low quality \cite{Park2019ICCV}.
Unfortunately, previous methods only work with eye region inputs, requiring high-quality images for training, and suffer from a lack of fidelity in preserving gaze in many cases.
We advance this task by generating high-fidelity face images with target gaze and head orientation along with control over many other independent factors.

\textbf{Cross-Dataset Gaze Estimation.}
Cross-dataset gaze estimation involves training and testing on different datasets and is a long-standing unsolved problem \cite{Zhang2019TPAMI}.
Shrivastava et al. propose to use a GAN for unsupervised domain adaptation of synthetic data \cite{Shrivastava2017CVPR}.
Wood\etal synthesize large numbers of eye images as training data for a k-NN model and test it on other datasets \cite{Wood2016ETRA}.
Representations of the internal geometry of the problem have helped with this too, with the 3D eyeball model used in \cite{Wang2018CVPR} or landmarks in \cite{Park2018ETRA}.
However, the lack of a means to model the distribution of the properties of a target test dataset remains a key challenge.
Our work is a meaningful step towards generating realistic images with controllable properties for cross-domain generic regression training.

\textbf{Fine-grained Conditional Image Generation.}
Conditional image generation takes as input an image and a condition, and generates a transformed image which reflects the condition while preserving other aspects of the input image. 
Many works achieve photo-realism in conditional image generation by 
carefully designing the flow of content and style information \cite{Huang2018ECCV,park2019semantic,lee2018diverse,mejjati2018unsupervised, configECCV2020}.
However, fine-grained control of conditions is more commonly performed by taking the target condition directly as input into a conditional GAN \cite{mirza2014conditional} as shown in StarGAN \cite{starGAN} and its derivatives \cite{he2019attgan,pumarola2018ganimation,wu2019relgan}.
For fine-grained control of continuous-valued conditions, transforming architectures, e.g., \cite{Hinton2011ICANN} are particularly effective. On synthetic data, transforming architectures are demonstrated to interpolate precisely in a measurable way \cite{Chen2019ICCV,mustikovelaCVPR20} or qualitatively \cite{Nguyen2019ICCV}, and show promising extrapolation possibilities \cite{Worrall2017ICCV}. These approaches are also effective in real-world downstream tasks such as semi-supervised human pose estimation \cite{Rhodin2018ECCV} and few-shot gaze estimator adaptation \cite{Park2019ICCV}, and thus are highly applicable to our task setting. 
We demonstrate a novel transforming architecture in this paper, which not only produces photo-realistic outputs despite noisy real-world training data, but also better reflects target gaze and head orientation conditions while disentangling extraneous factors.
\vspace{-1mm}
\section{Method}

\begin{figure}
    \centering
    \includegraphics[width=1.0\columnwidth]{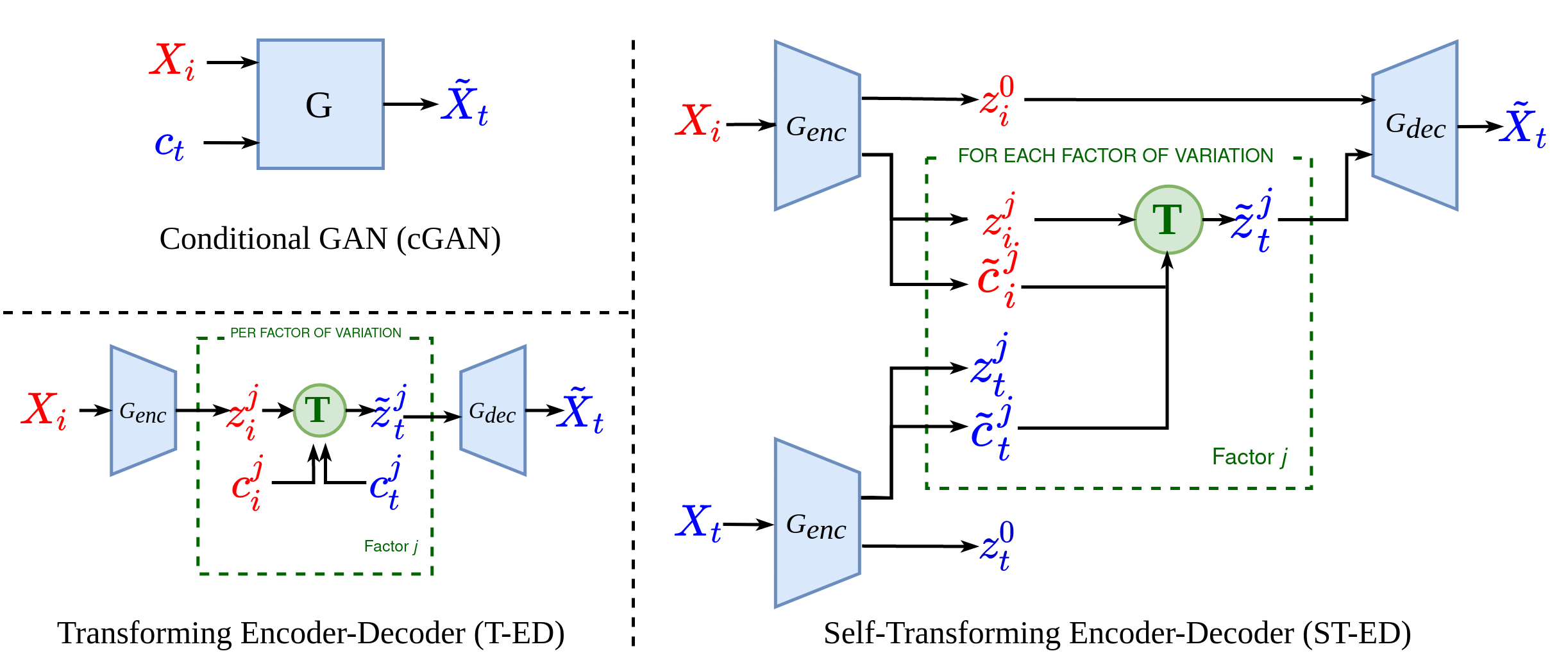}
    \caption{
    \textbf{Conceptual Overview.}
    Our proposed Self-Transforming Encoder-Decoders (ST-ED) can reduce reliance on noisy labels ($c_i$ and $c_t$) in the conditional image-to-image translation task: $\left(X_i, \boldsymbol{c_t}\right)\rightarrow X_t$.
    This is because the conditions used as input to the transformations are predicted as \emph{pseudo labels} ($\tilde{c}_i$ and $\tilde{c}_t$) and can deviate from the labels used for supervision.
    This also allows for extraneous factors to be learned without supervision on the predicted conditions.
    }
    \label{fig:overview}
\end{figure}

\subsection{Problem Setting}
Our goal is to train a conditional generative network, such that given an input image $X_i$ and a set of target conditions $\boldsymbol{c_t}$, it generates an output image $X_t$ by learning the mapping: $\left(X_i, \boldsymbol{c_t}\right)\rightarrow X_t$ (Fig.~\ref{fig:overview} top-left). Following previous works \cite{Worrall2017ICCV,Park2019ICCV}, we use a pair of images $X_i$ and $X_t$ of one person as input during training. We assume that each image can be represented by a personal non-varying embedding $z^0$ along with $N$ factors of image variations $\boldsymbol{f} = \{f^1, f^2, ..., f^{N}\}$. Each factor $f^j_i$ of an image $X_i$ can be described by an embedding $z_i^j$ and a continuous real-valued condition $c_i^j$, which describes the amount of deviation from the canonical state. Each condition $c_i^j$ is an $n$-dimensional vector where $n$ represents the degrees of freedom for the factor. Hence, $X_i$'s overall image condition is denoted by $\boldsymbol{c_i} = \{c_i^1, c_i^2, ..., c_i^{N}\}$. We further assume that a subset of the factors are ``extraneous" to the task ($\boldsymbol{f_u}$ with conditions $\boldsymbol{c_u}$) and are unknown \textit{a priori} (e.g., environmental lighting, hue, shadow, blurriness, and camera distance), while other ``explicit" factors are known (e.g., $c^g$ and $c^h$ for gaze and head orientation, respectively) and labeled with ground truth values. %

To solve this, we propose a novel architecture based on the concept of transforming encoder-decoders (T-ED), where labeled factors of image variation
are learned via rotationally equivariant mappings\footnote{This ``rotational equivariance'' is enforced for a factor by assuming that the apparent difference in rotations in two images as defined by their ground-truth labels must apply equivalently at the latent-code level. That is, if the ground-truth between samples $X_i$ and $X_t$ suggest that there is a gaze direction difference of $30^\circ$, this difference can be directly applied to the relevant latent code (via rotation matrices).} between independent latent embedding spaces and the image space
(Fig.~\ref{fig:overview} bottom-left). 
We extend T-ED to additionally discover, encode and disentangle multiple extraneous unknown factors in a self-supervised manner. We call our architecture the Self-Transforming Encoder-Decoder (ST-ED, Fig.~\ref{fig:overview} right). For a pair of images $X_i$ and $X_t$, ST-ED \emph{predicts} their personal non-varying codes ($z_i^0$ and $z_t^0$), their pseudo-label conditions ($\boldsymbol{\tilde{c}_i}$ and $\boldsymbol{\tilde{c}_t}$) and embedding representations ($\boldsymbol{z_i}$ and $\boldsymbol{z_t}$). Transforming with pseudo condition labels at training time, allows for the discovery of unknown extraneous factors in the absence of ground-truth.
In addition, pseudo-label predictions reduce reliance on noisy task-specific labels (typically used as input conditions in prior works) by performing the transformations with conditions that should better match the expected target image $X_t$. At inference time, we replace the pseudo target conditions $\boldsymbol{\tilde{c}_t}$ with the desired target conditions $\boldsymbol{c_t}$.

\subsection{Model overview}

Our redirection network $G$ consists of an encoder $G_{enc}$ and a decoder $G_{dec}$ (see Fig.~\ref{fig:overview} right). 
$G_{enc}$ estimates the factors $\boldsymbol{f}$ and the personal non-varying embedding $z^0$ of an image.
To begin the redirection process, we first encode both the input and target images by:
\begin{equation}
\{z_i^0, f_i^1, f_i^2, ..., f_i^{N}\} = G_{enc}(X_i), \quad
\{z_t^0, f_t^1, f_t^2, ..., f_t^{N}\} = G_{enc}(X_t), 
\end{equation}
where $f_i^{j} = \{z_i^{j}, \tilde{c}_i^{j}\}$ and $f_t^{j} = \{z_t^{j}, \tilde{c}_t^{j}\}$ are the predicted factors from the input and target images, respectively. We regard the condition $\tilde{c}^{j}$ predicted by $G_{enc}$ for the factor $f^{j}$ as its pseudo label.
Specifically in our setting, we define $f^g$ and $f^h$ as explicit factors corresponding to gaze and head orientation, and supervise their pseudo conditions with ground-truth labels $\boldsymbol{c}=\{c^g,c^h\}$ (see Sec. \ref{sec:objectives}).
We transform each input embedding $z_i^j$ with the pseudo label conditions $\tilde{c}_i^j$ and $\tilde{c}_t^j$ for the input and target images:
\begin{equation}
\tilde{z}^j_{t} = T(z^j_i, \tilde{c}^j_i, \tilde{c}^j_t)\\
=\boldsymbol{R}^j_t\left(\boldsymbol{R}^j_i \right)^{-1}z^j_i,\qquad j\in\{1, 2, ..., N\}
\end{equation}
where $\boldsymbol{R}^j_i$ and $\boldsymbol{R}^j_t$ are rotation matrices computed from $\tilde{c}^j_i$ and $\tilde{c}^j_t$, respectively. The transformation $T(\cdot)$ first rotates the input embedding $z^j_i$ back to a canonical representation and then to the target condition (see supplementary materials for more details).
Intuitively, the canonical orientation of gaze and head orientation would align with explicitly defined spherical coordinate systems (where the zero-configuration points towards the camera), whereas for extraneous factors such as directional lighting, the empirical mean observation of the training dataset may become associated with the canonical form.
The transformed embedding can be represented by $\boldsymbol{\tilde{z}_{t}} = \{z_i^0, \tilde{z}^1_{t}, \tilde{z}^2_{t}, ..., \tilde{z}^{N}_{t}\}$. Note that $z_t^0$ is directly passed through, assuming that for a given person (from whose data we sample $X_i$ and $X_t$) there exist unique non-varying factors that define their appearance.
Finally, we decode the transformed embeddings to generate the redirected image with $\tilde{X}_{t} = G_{dec}(\boldsymbol{\tilde{z}_{t}})$.

\subsection{Objectives}
\label{sec:objectives}

\textbf{Reconstruction Loss.} We guide the generation of redirected images with a pixel-wise $L_1$ reconstruction loss between the generated image $\tilde{X}_{t}$ and the target image $X_t$:
\begin{equation}
\mL{R}(\tilde{X}_{t}, X_t) =\frac{1}{|X_t|} \left\|\tilde{X}_{t} - X_t\right\|_1.
\end{equation}
\textbf{Functional Loss.} 
Inspired by the perceptual loss \cite{ He2019ICCV,johnson2016perceptual, Dosovitskiy2016}, we propose a novel functional loss which prioritizes the minimization of task-relevant inconsistencies between the generated and target images, e.g., the mismatch in iris positions for our case.
Analogous to its original form, we assess the generated image $\tilde{X}_t$ and target image $X_t$ with a feature-consistency loss which is formulated as the $L_2$ loss between the feature maps of $\tilde{X}_{t}$ and $X_t$:
\begin{equation}
\mL{F_{feature}}(\tilde{X}_{t}, X_t)=\sum_{i=1}^{5}\frac{1}{|\psi_i(X_t)|} \left\|\psi_i(\tilde{X}_{t}) - \psi_i(X_t)\right\|_2,
\end{equation}
where $\psi_i(\cdot)$ calculates the activation feature maps of the $i$-th layer of a network $F_d$, which is pre-trained on specific tasks (in our case, gaze direction and head orientation estimation) in order to extract task-specific features.
We additionally use a content-consistency loss which is formulated as the angular error between the predicted gaze direction and head orientation values from $\tilde{X}_{t}$ and $X_t$:
\begin{equation}
\mL{F_{content}}(\tilde{X}_{t}, X_t)
=
\mE{ang}\left(F_d^g\left(\tilde{X}_t\right), F_d^g\left(X_t\right)\right) + 
\mE{ang}\left(F_d^h\left(\tilde{X}_t\right), F_d^h\left(X_t\right)\right),
\end{equation}
\begin{equation}
\mE{ang}(\mathbf{v}, \hat{\mathbf{v}})=\arccos\cossim{\mathbf{v}}{\hat{\mathbf{v}}}.
\label{eq:angular_error}
\end{equation}

Note that instead of directly using the ground-truth labels of the target image, we assess the consistency of labels predicted by $F_d$. We hypothesize that this mitigates the deviation caused by the systematic error of $F_d$ and better aligns with our reconstruction objective.
Our final functional loss is defined as:
\begin{equation}
\mL{F} = \lambda_{F_{feature}}\mL{F_{feature}} + \mL{F_{content}}.
\end{equation}

\textbf{Disentanglement Loss.}
Individual factors should ideally be disentangled such that changing a subset of factors would not alter any of the other factors in the generated image $\tilde{X}_t$.
We encourage this disentanglement among the encoded factors, by first randomly transforming a subset of factors to create a mixed factor representation.
This is formulated as:
\begin{equation}
\boldsymbol{f_{mix}} = \{f^1_{mix}, f^2_{mix}, ..., f^N_{mix}\}, f^j_{mix} = s f^j_i + (1-s)\tilde{f}^j_{t}, \quad s \sim \{0,1\}.
\end{equation}
We decode these mixed embeddings and encode the synthesized image back to factors $\boldsymbol{f_{rec}} = G_{enc}(G_{dec}(\boldsymbol{z_{mix}}))$.
We represent the full disentanglement loss by the discrepancy between the mixed and recovered embeddings, $\boldsymbol{z_{mix}}$ and $\boldsymbol{z_{rec}}$, respectively, as well as by the error between the gaze and head labels before and after the process:
\begin{equation}
\mL{D} = \mE{ang}(\tilde{c}_{mix}^g, \tilde{c}_{rec}^{g}) +  \mE{ang}(\tilde{c}_{mix}^h, \tilde{c}_{rec}^{h})
+ \left(1 - \cossim{\boldsymbol{z_{mix}}}{\boldsymbol{z_{rec}}}\right).
\end{equation}
\textbf{Explicit Pseudo-Labels Loss.}
In our generator, two of our defined factors correspond to those for which we have explicit ground-truth labels, namely the gaze direction and head orientation.
Knowing this, the set of all pseudo labels $\boldsymbol{\tilde{c}}$ can be written as $\boldsymbol{\tilde{c}} = \{\tilde{c}^{u,1}, \tilde{c}^{u,2}, ..., \tilde{c}^{u,N-2}, \tilde{c}^{g}, \tilde{c}^{h}\}$, where $\tilde{c}^{u,j}$ represents the pseudo label for the $j$-th factor without available ground-truth (extraneous factors), and $\tilde{c}^{g}$ and $\tilde{c}^{h}$ are the pseudo labels for the gaze direction and head orientation (explicit factors). The corresponding ground-truth values are defined as $\{c^g,c^h\}$.
To guide the learning of the explicit factors, we use an estimation loss on the gaze direction and head orientation pseudo labels:
\begin{equation}
\mL{PL} = \mE{ang}(c^g, \tilde{c}^g) + \mE{ang}(c^h, \tilde{c}^h).
\end{equation}
\textbf{Embedding-Consistency Loss.}
As $X_i$ and $X_t$ are sampled from the same person's data, it is expected that the latent embeddings from different samples of the same person would be consistently learned.
However, as shown in \cite{Park2019ICCV}, a loss term for maintaining consistency helps in preserving person-specific information.
Therefore, we use a consistency loss between the transformed input embeddings $\boldsymbol{\tilde{z}_{t}}$ and the target embeddings $\boldsymbol{z_t}$, after flattening both embeddings:
\begin{equation}
\mL{EC} = 1 - \cossim{\boldsymbol{\tilde{z}_{t}}}{\boldsymbol{z_t}}.
\end{equation}
\textbf{GAN Loss.}
We use a standard GAN loss \cite{goodfellow2014generative} to encourage photo-realistic outputs from the generator $G$. We choose PatchGAN \cite{Isola2015pix2pix, Shrivastava2017CVPR} as our discriminator $D$ and use:
\begin{equation}
\mL{discriminator}(G,D) = \mathbb{E}\left[\log D\left(X_t\right) + \log\left(1- D\left(\tilde{X}_{t}\right)\right)\right],
\end{equation}
\begin{equation}
\mL{generator}(D) = \mathbb{E}\left[\log D\left(\tilde{X}_{t}\right)\right].
\end{equation}

\textbf{Full Loss.}
The combined loss function for the training of the generator is:
\begin{equation}
\mL{full} = \lambda_R \mL{R} + \lambda_F \mL{F} + \lambda_{PL} \mL{PL} + \lambda_{EC} \mL{EC} + \lambda_D \mL{D} + \mL{generator},
\end{equation}
where we empirically set $\lambda_R = 200, \lambda_F=20, \lambda_{PL}=5, \lambda_{EC} =2$, and $\lambda_D = 2$.

\section{Results}

\subsection{Implementation details}
We parameterize $G_{enc}$ and $G_{dec}$ with a DenseNet-based architecture as done in \cite{Park2019ICCV}.
We implement the external gaze direction and head orientation estimation network $F_d$ by a VGG-16 \cite{simonyan2014very} based architecture which outputs its predictions in spherical coordinates \cite{Zhang2019TPAMI}. For evaluation, we use a separate estimator $F^\prime_d$ network that is based on ResNet-50 \cite{ResNet} and is unseen during training, though trained on the same training data. %
We calculate ground truth labels $\{c^g,c^h\}$ via the data normalization procedure \cite{Sugano2014CVPR,Zhang2018ETRA} used to pre-process gaze datasets where head orientation is defined without the roll component.  %
Further implementation details are in our supplementary materials.

We train ST-ED using the GazeCapture training subset \cite{Krafka2016CVPR}, the largest publicly available gaze dataset. It consists of 1474 participants and over two million frames taken in unconstrained settings, which makes it challenging to train with. As such, to the best of our knowledge, we are the first to demonstrate that photo-realistic gaze redirection models can be learned from such noisy data.
To evaluate our models, we use GazeCapture (test subset), MPIIFaceGaze \cite{Zhang2017CVPRW}, Columbia Gaze \cite{Smith2013UIST}, and EYEDIAP \cite{FunesMora2014ETRA}. Each dataset exhibits different distributions of head orientation and gaze direction, as well as differences in the present extraneous factors.
This cross-dataset experiment allows for a better characterization of our approach in comparison to the state-of-the-art approaches.
\subsection{Evaluation Metrics}
\label{sec:evaluation_metrics}
We evaluate our gaze redirector with three kinds of measurements: redirection error, disentanglement of factors, and perceptual quality.
All metrics are better when lower in value.

\textbf{Redirection Error.}
We quantify the fulfillment of the explicit conditions in our image outputs by assessing gaze and head orientation with an external ResNet-50 based \cite{ResNet} estimator $F^\prime_d$ that is unseen during training.
We report the redirection error as the angular error between the estimated values from $F^\prime_d$ and their intended target values.

\textbf{Task Disentanglement Error.} 
In contrast to generalized disentanglement metrics such as the $\beta$-VAE metric \cite{higgins2017beta}, we directly measure the effect of other factors on the gaze and head orientation factors. 
To measure the effect of a factor $f^j$ on a factor $f^k, k\in \{g, h\}$, we first randomly perturb the condition $\tilde{c}^j$ with noise sampled from a uniform distribution,
$\varepsilon \sim \mathcal{U}\left(-\eta,\,\eta\right)$,
such that the perturbed condition is $\tilde{c}^{j^\prime} = \tilde{c}^j + \varepsilon$.
We choose $\eta = 0.1\pi$ in our evaluations. We then transform the associated embedding with $z_j^\prime = T(z_j, \tilde{c}_j, \tilde{c}_j^\prime)$ and decode the new embeddings via $\tilde{X}^\prime = G_{dec}(\boldsymbol{z^\prime})$. 
We apply the gaze and head orientation estimator $F^\prime_d$ on the \emph{perturbed} $\tilde{X}^\prime$ and unperturbed $\tilde{X}$ reconstructions.
We then compare the predicted gaze and head orientation directions from the two images using the error function $\mE{ang}$ (see Eq.~\ref{eq:angular_error}).
To perturb multiple factors, we simply take the average of all calculated errors over the number of factors perturbed.
We specifically evaluate
(a) $u\rightarrow g$, the effect of changes in all extraneous factors on apparent gaze direction,
(b) $u\rightarrow h$, the effect of changes in all extraneous factors on apparent head orientation, 
(c) $g\rightarrow h$, the effect of changes in the gaze direction factor on apparent head orientation and vice versa ((d) $h\rightarrow g$).

\textbf{LPIPS.}
The Learned Perceptual Image Patch Similarity (LPIPS) metric \cite{Zhang2018lpips} was previously used by \cite{He2019ICCV} to measure paired image similarity in gaze redirection, and thus we adopt it here.
Our ST-ED architecture is able to align to all predicted conditions from the target image (\emph{all} in Tab.~\ref{tab:ablation}) or just the gaze direction and head orientation conditions (\emph{$g+h$} in Tab.~\ref{tab:ablation}) and as such we report both scores.

\begin{table}
    \centering
    \caption{
        \textbf{Ablation study (lower is better).}
        Our FAZE~\cite{Park2019ICCV}-like T-ED base model learns only explicit factors (gaze direction and head orientation). We found that
        controlling these explicit factors with self-predicted pseudo labels (ST-ED Base Model) helps to improve redirection and disentanglement scores. Building on the ST-ED architecture, we allow for extraneous factors $\boldsymbol{f_u}$ to be discovered which further boosts performance. Additionally, our novel functional ($\mL{F}$) and disentanglement ($\mL{D}$) losses enforce accurate generation of task-relevant features and clean disentanglement between different factors, improving all metrics.
        Each of the changes listed in the first column are with respect to the immediate previous row.
        \label{tab:ablation}
    }
    \begin{tabular}{l|ccc|ccc|cc}
        \hline
        \multirow{2}{*}{Approach} & 
        \multicolumn{3}{c|}{Gaze Direction} & 
         \multicolumn{3}{c|}{Head Orientation} &
        \multicolumn{2}{c}{LPIPS} \\
        \cline{2-9}
        & Re-dir. & $u \rightarrow g$ &  $h\rightarrow g$ 
        & Re-dir. &  $u \rightarrow h$  &  $g\rightarrow h$ 
        & $g + h$ & all \\
        \hline
        T-ED Base Model \cite{Park2019ICCV}   & 7.114& - &4.882 & 2.470& -  & 0.542&0.279& 0.279 \\
        ST-ED Model   & 5.107& - &3.639 &1.479 & -  & 0.660&0.272& 0.271 \\
        $ + \boldsymbol{f_u}$ &4.716 & 0.814& 3.404& 1.434& 0.314& \textbf{0.385}&0.257& 0.215\\
        $ + \mL{F} + \mL{D}$
        &\textbf{2.195} &\textbf{0.507} &\textbf{2.072} &\textbf{0.816} &\textbf{0.211} &0.388 &\textbf{0.248}& \textbf{0.205} \\
        \hline
    \end{tabular}
\end{table}
\blockcomment{
\vspace{-2mm}
\begin{figure}
\centering
\begin{subfigure}[b]{.48\linewidth}
\includegraphics[width=\linewidth]{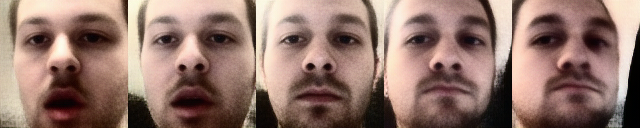}
\vskip -1mm
\caption{without $\mL{D}$}
\label{fig:wo/D}
\end{subfigure}
\hfill
\begin{subfigure}[b]{.48\linewidth}
\includegraphics[width=\linewidth]{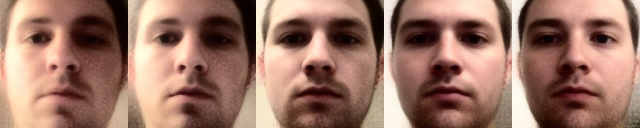}
\vskip -1mm
\caption{with $\mL{D}$}
\end{subfigure}
\vspace{-2mm}
\caption{Comparison of training with (b) and without (a) the disentanglement loss term $\mL{D}$ which enforces disentanglement between task-independent extraneous factors and task-dependent explicit factors (gaze direction and head orientation). Changing only the extraneous factors can still affect the reconstructed head orientation (a), and training with $\mL{D}$ prevents such entanglement.
}
\label{fig:example_unsupervised_redirection}
\vspace{-5mm}
\end{figure}
}
\subsection{Ablation study}
We report results of an ablation study evaluated on the GazeCapture test set \cite{Krafka2016CVPR} in Table \ref{tab:ablation}, where we show the three metrics.
Our base model uses a transforming encoder-decoder (T-ED) architecture with only gaze direction and head orientation factors, which is similar to FAZE \cite{Park2019ICCV}. We train our T-ED base model with reconstruction $\mL{R}$, adversarial  $\mL{generator}$ and embedding consistency  $\mL{EC}$ losses.
Our ST-ED model in the second row of Table \ref{tab:ablation} additionally predicts gaze and head pseudo labels which are used for transforming the corresponding factors. We can see that controlling the explicit transformations with self-predicted pseudo labels during training helps to improve the redirection and disentanglement scores. This is because self-prediction of pseudo-labels helps to reduce the harm of confounding and noisy labels.
From the third row of Table \ref{tab:ablation}, we can see that learning to discover and disentangle the extraneous factors via our framework improves redirection, disentanglement and LPIPS scores further. When using only explicit, task-specific factors, the model is forced to embed extraneous changes into the gaze and head orientation factors in order to satisfy reconstruction-related losses, thus deteriorating redirection accuracy in outputs.
We also see that when the extraneous factors are aligned to predicted conditions in the target image, the LPIPS score improves (all vs $g+h$).
Qualitatively, we can see in Fig.~\ref{fig:s+u} that our approach can align better to a given target image on additionally transforming our extraneous factors, compared to when only redirecting the explicit factors (Fig.~\ref{fig:s}).
When we add the functional $\mL{F}$ and disentanglement $\mL{D}$ losses (row 4), they yield large and consistent improvements in all error metrics. The functional loss encourages accurate reconstruction of task-relevant features, while the disentanglement loss enforces that explicit variations are only influenced by the corresponding factors of ST-ED. Our results indicate that the disentanglement of factors and the reconstruction of task-related conditions are aligned objectives.

\subsection{Comparisons to the state-of-the-art baselines}
\begin{table}[h!]
    \centering
    \caption{
        \textbf{State-of-the-art comparisons.}
        We compare our best model against StarGAN \cite{starGAN} and He\etal \cite{He2019ICCV} on the task of full-face gaze and head redirection, evaluated on four gaze datasets.
        Our approach not only generates gaze direction and head orientation more faithfully, but also achieves better disentanglement for separetely controlling the two properties.
        Furthermore, our model allows for the manipulation of extraneous factors, enabling us to out-perform in terms of perceptual image quality as well (for the row \emph{Ours}, we calculate LPIPS after aligning all factors to a target image using its pseudo-labels $\boldsymbol{\tilde{c}_t}$).
    }
    \label{tab:sota}
    \footnotesize
    \renewcommand{\arraystretch}{1.0}
    \setlength\tabcolsep{1mm}
    \begin{subtable}{0.46\columnwidth}
        \centering
        \caption{GazeCapture}\vskip -2mm
        \begin{tabular}{l|ccccc}
            \hline
            & \makecell[c]{Gaze\\[-0.5mm]Redir.} & \makecell[c]{Head\\[-0.5mm]Redir.} & $g\rightarrow h$ & $h\rightarrow g$ & LPIPS \\
            \hline
            StarGAN & 4.602& 3.989&0.755 &3.067 &0.257 \\
            He\etal &4.617 & 1.392 & 0.560 & 3.925 & 0.223 \\
            Ours & \textbf{ 2.195}& \textbf{ 0.816}&\textbf{ 0.388} &\textbf{ 2.072}& \textbf{0.205}\\
            \hline
        \end{tabular}
    \end{subtable}
    \hfill
    \begin{subtable}{0.46\columnwidth}
        \centering
        \caption{MPIIFaceGaze}\vskip -2mm
        \begin{tabular}{l|ccccc}
            \hline
            & \makecell[c]{Gaze\\[-0.5mm]Redir.} & \makecell[c]{Head\\[-0.5mm]Redir.} & $g\rightarrow h$ & $h\rightarrow g$ & LPIPS \\
            \hline
            StarGAN & 4.488&  3.031& 0.786 &  2.783& 0.260\\
            He\etal &  5.092&1.372 & 0.684 &3.411& 0.241\\
            Ours & \textbf{ 2.233}& \textbf{ 0.884}&\textbf{ 0.365}&\textbf{ 1.849} & \textbf{0.203}\\
            \hline
        \end{tabular}
    \end{subtable}
    
    \begin{subtable}{0.46\columnwidth}
        \centering
        \caption{Columbia}\vskip -2mm
        \begin{tabular}{l|ccccc}
            \hline
            & \makecell[c]{Gaze\\[-0.5mm]Redir.} & \makecell[c]{Head\\[-0.5mm]Redir.} & $g\rightarrow h$ & $h\rightarrow g$ & LPIPS \\
            \hline
            StarGAN &6.522&3.444&1.029& 3.359&0.255 \\
            He\etal &7.345 & 1.677 & 0.692& 3.831& \textbf{0.227} \\
            Ours & \textbf{ 3.333}& \textbf{ 1.095}&\textbf{ 0.452}&\textbf{2.136} & 0.242\\
            
            \hline
        \end{tabular}
    \end{subtable}
    \hfill
    \begin{subtable}{0.46\columnwidth}
        \centering
        \caption{EYEDIAP}\vskip -2mm
        \begin{tabular}{l|ccccc}
            \hline
            & \makecell[c]{Gaze\\[-0.5mm]Redir.} & \makecell[c]{Head\\[-0.5mm]Redir.} & $g\rightarrow h$ & $h\rightarrow g$ & LPIPS \\
            \hline
            StarGAN &14.906&3.929& 0.915 &4.025&0.248\\
            He\etal  &13.548 &1.581 & 0.663 &  4.367 &0.218\\
            Ours & \textbf{11.290}&\textbf{ 0.919}&\textbf{ 0.402} &\textbf{2.670}& \textbf{0.213}\\
            \hline
        \end{tabular}
    \end{subtable}
\end{table}

\begin{figure}[t]
\centering
\begin{subfigure}{.15\linewidth}
\includegraphics[width=\linewidth]{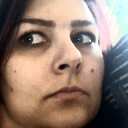}
\end{subfigure}
\begin{subfigure}{.15\linewidth}
\includegraphics[width=\linewidth]{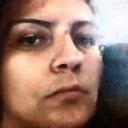}
\end{subfigure}
\begin{subfigure}{.15\linewidth}
\includegraphics[width=\linewidth]{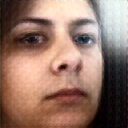}
\end{subfigure}
\begin{subfigure}{.15\linewidth}
\includegraphics[width=\linewidth]{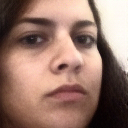}
\end{subfigure}
\begin{subfigure}{.15\linewidth}
\includegraphics[width=\linewidth]{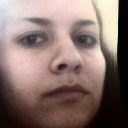}
\end{subfigure}
\begin{subfigure}{.15\linewidth}
\includegraphics[width=\linewidth]{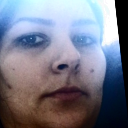}
\end{subfigure}
\begin{subfigure}{.15\linewidth}
\includegraphics[width=\linewidth]{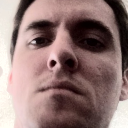}
\end{subfigure}
\begin{subfigure}{.15\linewidth}
\includegraphics[width=\linewidth]{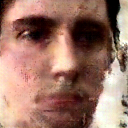}
\end{subfigure}
\begin{subfigure}{.15\linewidth}
\includegraphics[width=\linewidth]{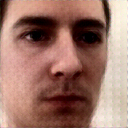}
\end{subfigure}
\begin{subfigure}{.15\linewidth}
\includegraphics[width=\linewidth]{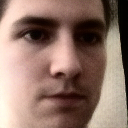}
\end{subfigure}
\begin{subfigure}{.15\linewidth}
\includegraphics[width=\linewidth]{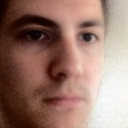}
\end{subfigure}
\begin{subfigure}{.15\linewidth}
\includegraphics[width=\linewidth]{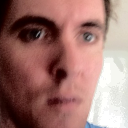}
\end{subfigure}
\\
\begin{subfigure}{.15\linewidth}
\includegraphics[width=\linewidth]{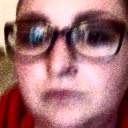}
\end{subfigure}
\begin{subfigure}{.15\linewidth}
\includegraphics[width=\linewidth]{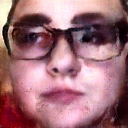}
\end{subfigure}
\begin{subfigure}{.15\linewidth}
\includegraphics[width=\linewidth]{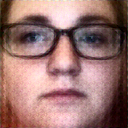}
\end{subfigure}
\begin{subfigure}{.15\linewidth}
\includegraphics[width=\linewidth]{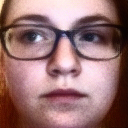}
\end{subfigure}
\begin{subfigure}{.15\linewidth}
\includegraphics[width=\linewidth]{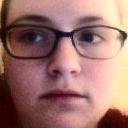}
\end{subfigure}
\begin{subfigure}{.15\linewidth}
\includegraphics[width=\linewidth]{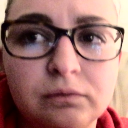}
\end{subfigure}
\\
\begin{subfigure}{.15\linewidth}
\includegraphics[width=\linewidth]{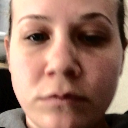}
\end{subfigure}
\begin{subfigure}{.15\linewidth}
\includegraphics[width=\linewidth]{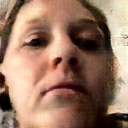}
\end{subfigure}
\begin{subfigure}{.15\linewidth}
\includegraphics[width=\linewidth]{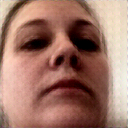}
\end{subfigure}
\begin{subfigure}{.15\linewidth}
\includegraphics[width=\linewidth]{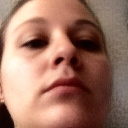}
\end{subfigure}
\begin{subfigure}{.15\linewidth}
\includegraphics[width=\linewidth]{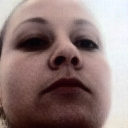}
\end{subfigure}
\begin{subfigure}{.15\linewidth}
\includegraphics[width=\linewidth]{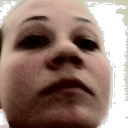}
\end{subfigure}
\\
\subcaptionbox{Input image}[.15\linewidth]{\includegraphics[width=\linewidth]{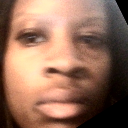}}
\subcaptionbox{StarGAN}[.15\linewidth]{\includegraphics[width=\linewidth]{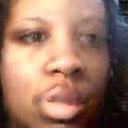}}
\subcaptionbox{He et al.}[.15\linewidth]{\includegraphics[width=\linewidth]{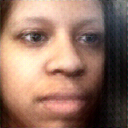}}
\subcaptionbox{Ours ($g+h$)
\label{fig:s}}[.15\linewidth]{\includegraphics[width=\linewidth]{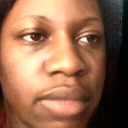}}
\subcaptionbox{Ours (all)
\label{fig:s+u}}[.15\linewidth]{\includegraphics[width=\linewidth]{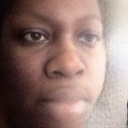}}
\subcaptionbox{Target Image}[.15\linewidth]{\includegraphics[width=\linewidth]{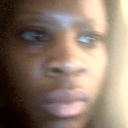}}

\caption{
    \textbf{Qualitative Results.}
    Our method produces more detailed and photo-realistic images compared to the baseline methods of He\etal \cite{He2019ICCV} and StarGAN \cite{starGAN}.
    By aligning to all predicted pseudo-labels of a target ground-truth image, our approach can also better approximate the target.
}
\label{fig:qualitative_main}
\vspace{-2mm}
\end{figure}
We compare against He\etal \cite{He2019ICCV} and StarGAN \cite{starGAN} by adapting their approaches to our setting of using face input images and allowing for continuous value gaze and head orientation redirection. He\etal \cite{He2019ICCV} is the state-of-the-art in gaze redirection methods and is more accurate and photorealistic than previous approaches such as DeepWarp \cite{Ganin2016ECCV} and does not require synthetic training data as in Yu\etal \cite{Yu2019CVPR}. StarGAN \cite{starGAN} is similar in approach to He\etal but is a more generic approach and thus important to compare to. We re-implement all baselines with a DenseNet-based architecture similar to ours to yield comparable results (implementation details are in the supplementary).

Our method outperforms the baseline methods, as measured by both the redirection and disentanglement metrics on four different datasets (see Tab.~\ref{tab:sota}). We achieve precise and stricter control over gaze and head orientation thanks to the clear separation of task-relevant explicit factors and self-learned extraneous factors. The latter encode task-irrelevant changes which are disentangled from the explicit factors.
As can be seen from Fig. \ref{fig:qualitative_main}, our method can generate photo-realistic results even for subjects with glasses and under large head orientation changes.  
He\etal \cite{He2019ICCV} show periodic artifacts on faces and backgrounds. This may be due to the fact that only a perceptual loss is used as the reconstruction objective (instead of a pixel-wise L1/L2 loss), resulting in a constraint that is applied at lower image scales than the original resolution, which leads to such artifacts. 
StarGAN \cite{starGAN} maintains personal appearances well under small head orientation changes, but degenerates quickly with larger redirection angles. These comparisons show the importance of aligning extraneous factors between images during training which allows for stricter supervision on the generated images. Furthermore, our method can optionally match the lighting conditions of the target image by transforming the extraneous latent factors, as can be seen from columns (d) and (e).
Importantly, our method more faithfully reproduces the target gaze direction and head orientation as shown in both qualitative and quantitative results. For more qualitative results, please refer to the supplementary materials.

\subsection{Semi-supervised Cross-dataset Gaze Estimation}
The value of learning to robustly isolate and control explicit factors from in-the-wild data lies in its potential to improve performance of downstream computer vision tasks, such as in training gaze or head orientation estimation models. %
Therefore, we perform experiments on semi-supervised person-independent cross-dataset estimation on four popular gaze datasets.
We show that even with small amounts of training data, %
our gaze redirector can extract and understand the variation of the dataset's factors sufficiently to augment it with new samples without introducing errors.

We randomly select $x \in\{2.5k, 5k, 10k, 20k, 50k\}$ labeled samples from the GazeCapture~\cite{Krafka2016CVPR} training split and use the rest \emph{without} labels to train ST-ED. While training, we employ pseudo label estimation loss $\mL{PL}$ only for labeled images, and the $F_d$ used for functional loss $\mL{F}$ is also pretrained on only the labeled subset. 
In forming our batches, we select only labeled samples for $X_i$ while $X_t$ can be either a labeled or unlabeled sample from the same person as $X_i$.

Once ST-ED is trained, we estimate the joint probability distribution function of the gaze and head orientation values of the labeled subset and sample random target conditions from it. We redirect the images in the labeled subset to these target conditions using ST-ED, and add them to the labeled set to create an augmented dataset (that is 2x the size of the labeled set). Lastly, we train a new gaze and head orientation estimation network (with the same architecture as $F_d$), but with this augmented set and compare its performance to the ``baseline" version trained only with the smaller labeled dataset.

\begin{figure}[t!]
    \centering
    \includegraphics[width=0.235\columnwidth]{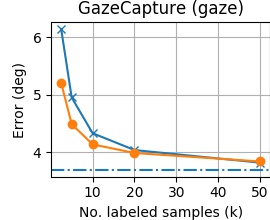}
    \includegraphics[width=0.235\columnwidth]{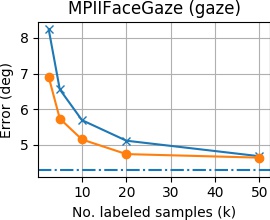}
    \includegraphics[width=0.235\columnwidth]{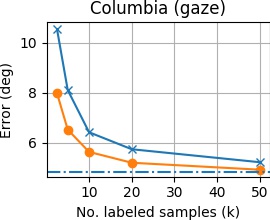}
    \includegraphics[width=0.235\columnwidth]{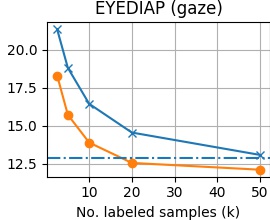}
    \\[2mm]
    \includegraphics[width=0.235\columnwidth]{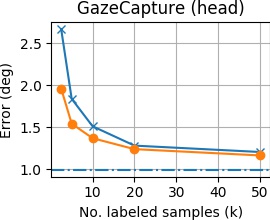}
    \includegraphics[width=0.235\columnwidth]{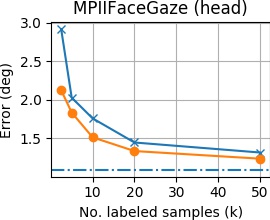}
    \includegraphics[width=0.235\columnwidth]{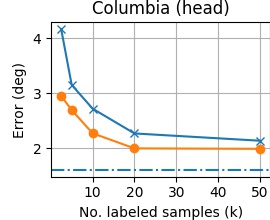}
    \includegraphics[width=0.235\columnwidth]{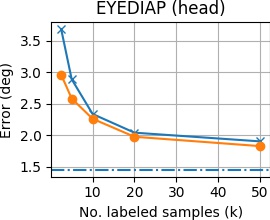}
    \\[2mm]
    \includegraphics[width=0.5\columnwidth]{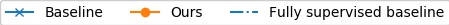}
    \\[-1mm]
    \caption{
        \textbf{Augmenting labeled training data via semi-supervised ST-ED.}
        We show that a semi-supervised learned ST-ED can augment an initial labeled dataset via redirection, to yield consistent improvements in cross-dataset gaze estimation benchmarks.
        Our method is consistently better than the baseline estimator trained on the initial labeled data only. This holds true for both gaze direction and head orientation estimation. 
    }
    \label{fig:semi-supervise-plot}
\end{figure}

Fig.~\ref{fig:semi-supervise-plot} shows that the gaze and head orientation estimation networks trained with both labeled data and augmented data (via redirection with a semi-supervised ST-ED) yields consistently improved performance on all four evaluation datasets.
This is particularly true for cases with very few labeled samples (2,500), where the largest gains in performance are found.
Approximately speaking, our method requires half the amount of labeled data to achieve similar performance as the baseline model.
Importantly in our setting, our redirection network is not trained on any additional labeled data. %
This is in contrast to prior works which use additional synthetic \cite{Yu2019CVPR} or real-world data \cite{He2019ICCV} in the training of the redirector. Hence, this setting is very challenging, and to the best of our knowledge, we are the first to tackle this problem, demonstrating consistent improvements in downstream performance on multiple real-world datasets.

\section{Discussion}
We have discussed our method in the context of gaze and head-orientation estimation. However, we note that the architecture is motivated in a general sense and thus we are optimistic in its potential application to other conditional image generation tasks for which a small subset of labels is available and many more factors of variation need to be identified and separated without explicit supervision. We can discover and match the mis-alignment of extraneous variations with our self-learned transformations, and thus improve the learning of task-relevant factors. Furthermore, our proposed functional loss punishes perceptual differences between images with an emphasis on task-relevant features, which can be useful for various problems with an image reconstruction objective, for e.g., auto-encoding, neural rendering, etc. We leave further exploration of different application domains for future work.

\clearpage

\section*{Broader Impact}
\vspace{-3mm}
Our work can perform accurate and photo-realistic gaze and head orientation redirection which makes augmenting existing datasets possible. It can also be used for film post-editing, group photo editing and video conferencing to correct the gaze directions and head orientations. We believe the method may have applications in other problem settings and thus it should be possible to leverage it to generate training data for estimators beyond gaze. 
Given that the method can generate realistic looking images under fine-grained control of selected parameters, it could also be leveraged for malicious manipulation of imagery in the context of ``deep-fakes''. Due to the limitations of currently available gaze datasets, our method does not yet handle extreme gaze directions that are beyond the distribution of the training dataset, nor fully faithfully preserve person-specific details in its redirection output. However, future developments should keep the ethical and privacy concerns in mind when refining such technologies.

\begin{ack}

This project has received funding from the European Research Council (ERC) under the European Union’s Horizon 2020 research and innovation programme grant agreement No. StG-2016-717054.

\begin{figure}[h!]
    \centering
    \includegraphics[width=0.4\columnwidth]{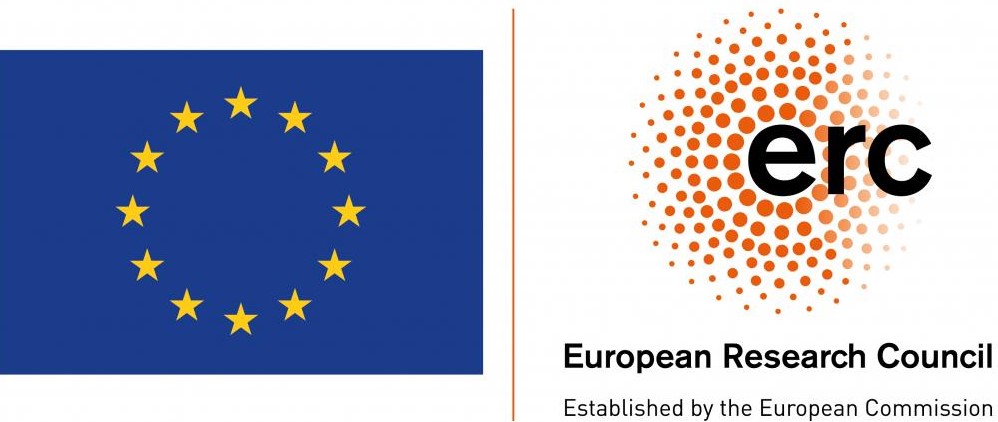}
\end{figure}

\end{ack}

\clearpage
\begin{center}
\textbf{\large Supplemental Materials}
\end{center}
\setcounter{equation}{0}
\setcounter{figure}{0}
\setcounter{table}{0}
\setcounter{page}{1}
\setcounter{section}{0}

\section{Overview}
In this supplementary document, %
we first show additional qualitative examples and experimental results (Sec. \ref{sec:results}).
We then provide details of the transformation function $T(\cdot)$ (Sec. \ref{sec:transformation}) and implementation of network architectures (Sec. \ref{sec:Implementation}). We highly recommend that our readers view the supplementary video which provides further results produced by the proposed method.

\section{Further Results}
\label{sec:results}

\subsection{Video Samples}
Please check the accompanying video for samples which further demonstrate the quality and consistency of our approach. Note that all samples are produced using a test subset of the GazeCapture dataset \cite{Park2019ICCV} and as such no over-fitted results are shown.

Our ST-ED approach is able to reconstruct smoother and more plausible changes in gaze direction and head orientation, and generates images with photo-realism despite being trained on a highly noisy dataset of in-the-wild images. Furthermore, the gaze direction and head orientation apparent in the output video sequences more faithfully reflect the given inputs, with promising results at extreme angles which go beyond the range of the training dataset (as such, the faithfulness of those generated samples cannot be quantitatively measured yet).

\subsection{Additional Image Samples}
In Fig. \ref{fig:additional qualitative results} (see end of document), we show additional qualitative results from the GazeCapture test set, comparing our method against state-of-the-art baselines. 

\subsection{Failure Cases}
\label{sec:failure_case}
In Fig.~\ref{fig:failures} we show that our method sometimes exhibits difficulties in handling eyeglasses and expressions (Fig. \ref{fig:failure_glasses}), preserving person-specific appearance characteristics such as face shape (Fig.~\ref{fig:failure_shape}), or retaining finer details of the face such as moles and freckles (Fig.~\ref{fig:failure_freckles}).

\begin{figure}[!htb]
\centering
\begin{subfigure}[t]{0.31\columnwidth}
    \centering\begin{tabular}{@{}c@{\hspace{2mm}}c@{}}
        \includegraphics[width=0.48\linewidth]{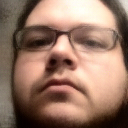} &
        \includegraphics[width=0.48\linewidth]{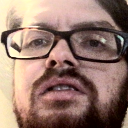} \\
        \footnotesize Generated & \footnotesize Target \\[-1mm]
    \end{tabular}
    \caption{Eyeglasses and Expressions\label{fig:failure_glasses}}
\end{subfigure}
\hfill
\begin{subfigure}[t]{0.31\columnwidth}
    \centering\begin{tabular}{@{}c@{\hspace{2mm}}c@{}}
        \includegraphics[width=0.48\linewidth]{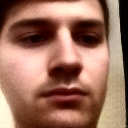} &
        \includegraphics[width=0.48\linewidth]{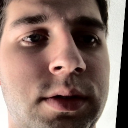} \\
        \footnotesize Generated & \footnotesize Target \\[-1mm]
    \end{tabular}
    \caption{Person-specific face shape\label{fig:failure_shape}}
\end{subfigure}
\hfill
\begin{subfigure}[t]{0.31\columnwidth}
    \centering\begin{tabular}{@{}c@{\hspace{2mm}}c@{}}
        \includegraphics[width=0.48\linewidth]{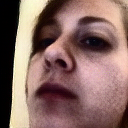} &
        \includegraphics[width=0.48\linewidth]{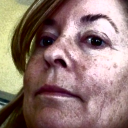} \\
        \footnotesize Generated & \footnotesize Target \\[-1mm]
    \end{tabular}
    \caption{Finer details (freckles)\label{fig:failure_freckles}}
\end{subfigure}
\caption{
    \textbf{Failure cases and limitations.}
    Most of our failure cases are related to appearance preservation. Difficult scenarios includes eyeglasses, unusual facial features or expressions and finer details such as freckles.
    \label{fig:failures}
}
\end{figure}

\subsection{Experiments on loss weight combinations}
In Table \ref{weights}, we show that our method is robust to different wights for the various loss terms. The fluctuation in redirection score is around 0.1 degrees when decreasing reconstruction, functional or pseudo label loss weights by half (compared to the final weights that we selected and show in Sec.~3.3 of the main paper). We also experiment with varying the weights of our embedding loss terms, which includes the embedding consistency loss $\mL{EC}$ and the embedding part of our disentanglement loss $\mL{D}$ (the last term in Eq.~9 of the main paper). Removing the embedding-related loss terms leads to slightly better LPIPS metric but worsened gaze redirection metric. Overall, we conclude that our loss combination is stable and robust to different weight values.
\begin{table}
    \centering
    \caption{
        \textbf{Sensitivity to loss term weights.}\label{weights} Our method is robust to different loss term weight combinations. The metrics remain reasonably consistent when the weight of a loss term is decreased by half or set to zero (for embedding losses).
    }
   \begin{tabular}{l|ccc|ccc|cc}
        \hline
        \multirow{2}{*}{Approach} & 
        \multicolumn{3}{c|}{Gaze Direction} & 
         \multicolumn{3}{c|}{Head Orientation} &
        \multicolumn{2}{c}{LPIPS} \\
        \cline{2-9}
        & Re-dir. & $u \rightarrow g$ &  $h\rightarrow g$ 
        & Re-dir. &  $u \rightarrow h$  &  $g\rightarrow h$ 
        & $g + h$ & all \\
        \hline
        $0.5$x weight for $\mL{R}$ & 2.231& 0.517 &\textbf{2.003} &\textbf{ 0.814}& 0.209 & 0.390 &0.247& 0.204 \\
        $0.5$x weight for $\mL{F}$   & 2.321& 0.558 &2.225 &0.831 & 0.227 &0.397&0.253& 0.206 \\
        $0.5$x weight for $\mL{PL}$ &2.249 & 0.550& 2.070& 0.819& 0.229& 0.411&0.248& 0.204\\
        No embedding losses &2.359 & \textbf{0.470}& 2.013&0.779&\textbf{0.198}& 0.404&\textbf{0.243}& \textbf{0.200}\\
        Ours &\textbf{2.195} &0.507 &2.072 & 0.816 &0.211 &0.388 &0.248& 0.204 \\
        \hline
    \end{tabular}
\end{table}

\subsection{Comparison with He\etal \cite{He2019ICCV} on the Downstream Task}

\begin{table}
    \centering
    \caption{
        Downstream estimation error with $2.5k$ real training samples. We show the results from He\etal \cite{He2019ICCV}, a supervised baseline method and our method for both gaze and head pose estimation tasks. The redirector is trained on the whole GazeCapture training set for He\etal \cite{He2019ICCV}, and only $2.5k$ real samples for our method (in a semi-supervised fashion with unlabeled samples from the rest of the dataset).
    }
    \label{tab:he_etal_aug}
    \footnotesize
    \vskip -2mm
    \begin{subtable}{\columnwidth}
        \centering
        \caption{Gaze Direction}
        \vskip -2mm
        \begin{tabular}{l|ccccc}
            \hline
            Method & GazeCapture & MPIIGaze & ColumbiaGaze & EYEDIAP \\
            \hline
            He\etal \cite{He2019ICCV} & 9.882 & 11.985 & 9.274 & 25.43 \\
            Baseline & 6.138 & 8.243 &  10.536 & 21.35 \\
            Ours & \textbf{5.203} & \textbf{6.903} & \textbf{7.974} & \textbf{18.31} \\
            \hline
        \end{tabular}
    \end{subtable}
    \\[0mm]
    \begin{subtable}{\columnwidth}
        \centering
        \caption{Head Orientation}
        \vskip -2mm
        \begin{tabular}{l|ccccc}
                       \hline
            Method & GazeCapture & MPIIGaze & ColumbiaGaze & EYEDIAP \\
            \hline
            He\etal \cite{He2019ICCV} & 13.457 & 7.964 & 3.787 & 13.04 \\
            Baseline & 2.668 & 2.913 & 4.158 & 3.681  \\
            Ours & \textbf{1.961} & \textbf{2.122} & \textbf{2.950} & \textbf{2.960} \\
            \hline
        \end{tabular}
    \end{subtable}
\end{table}

We compare our method with baseline methods on the tasks of semi-supervised cross-dataset gaze and head pose estimation in Sec.~4.5 of the main paper. In this section, we further compare it with the previous state-of-the-art method from He\etal \cite{He2019ICCV}. We first train the gaze redirector of the method from He\etal \cite{He2019ICCV} with the \emph{entire} GazeCapture training set, and then generate eye images from $2.5k$ real samples. Finally we train the estimator $F_d$ with both the real and generated data.
Please note that our method as well as the baseline follow the same procedure where only the $2.5k$ real samples are used during the entire training procedure.

As shown in Table~\ref{tab:he_etal_aug}, augmenting the real data samples using the method from He\etal \cite{He2019ICCV} generally results in performance degradation compared to the baseline, despite having used more labeled data for training the redirection network. This difference in implementation was necessary, as with very few samples the redirection network of He\etal could not be trained successfully. Furthermore, the approach of He\etal \cite{He2019ICCV} cannot be trained in a semi-supervised manner.

This result highlights that smaller differences in redirector performance can cause large differences in downstream regression tasks. We believe that the degradation in performance is also due to the lower quality of the generated images from He\etal \cite{He2019ICCV}, which contain many artifacts as shown in Fig. \ref{fig:additional qualitative results}. This causes a domain shift problem between the training images (half of which are generated images) and the testing images (real), which harms performance.  

In contrast, our semi-supervised re-director is trained without any additional labeled data. Nonetheless, it can generate accurate and photo-realistic samples, which consistently improves performance over the baseline estimator.

\section{Definition of Transformations}
\label{sec:transformation}

In a typical transforming encoder-decoder architecture, the encoder predicts an embedding and this embedding is transformed via pre-defined transformation routines such as translations as shown in the initial transforming autoencoder architecture by Hinton\etal \cite{Hinton2011ICANN}. We follow the approach of Worrall\etal \cite{Worrall2017ICCV} and Park\etal \cite{Park2019ICCV} and define our transformations as rotations, which are easily invertible and are linear orthogonal transformations. This makes such transformations easy to control and to some extent, interpret. 

For a given factor of variation $f^j_i$, we assume as described in Sec.~3.2 of the main paper that this factor is described by an embedding $z^j_i$ and a pseudo condition $\tilde{c}^j_i$. More specifically, assuming that a rotation matrix $\boldsymbol{R}^j_i$ is associated with this factor and its variation, we define that the embedding predicted by the encoder $G_{enc}$ is written as,
\begin{equation}
    z^j_i=\boldsymbol{R}^j_i z^j_{i_{canon}},
\end{equation}
where $z^j_{i_{canon}}$ is the canonical representation associated with input $X_i$.
Based on this assumption, we have:
\begin{align}
    \tilde{z}^j_t 
    &= T\left(z^j_i,\,c^j_i,\,c^j_t\right) \\
    &=
        \boldsymbol{R}^j_t
        \left(
        \boldsymbol{R}^j_i
        \right)^{-1}
        z^j_i,
\end{align}
where the rotation matrices $\boldsymbol{R}^j_t$ and $\boldsymbol{R}^j_i$ are computed from $\tilde{c}^j_t$ and $\tilde{c}^j_i$, respectively.
By the definition of $SO(\cdot)$ rotation matrices, the inverse of a given matrix is simply its transpose.

More specifically, at the stage of configuring the ST-ED architecture (introduced in Sec.~3.1 of the main paper), an arbitrary number of factors can be defined with $\boldsymbol{f}=\left\{ f^1,\,f^2,\,\ldots,\,f^N\right\}$ where each factor $f^j$ can be controlled with degrees of freedom $\in\left\{0,\,1,\,2\right\}$. 

\textbf{0-DoF Factors.} 
For the case of zero degrees of freedom, we define $z^0_i$, which does not vary but is simply passed to the decoder assuming and enforcing that $z^0_i\simeq z^0_t$ via the reconstruction objective (Eq. 3 of main paper).

\textbf{1-DoF Factors.} 
For the case of $1$ degree of freedom, we define the rotation matrix of factor $f^j_i$ as:
\begin{equation}
    \boldsymbol{R}^j_i = \begin{pmatrix}
    \cos{c^j_i} & -\sin{c^j_i} \\
    \sin{c^j_i} &  \cos{c^j_i} \\
    \end{pmatrix}
\end{equation}
and the dimensionality of the associated embedding $z^j_i$ becomes $N_f^j\times2$, where $N^j_f$ is the hyperparameter for defining the latent embedding size for this 1-dimensional factor.

\textbf{2-DoF Factors.} 
For the case of $2$ degrees of freedom, we define the rotation matrix of factor $f^j_i$ as:
\begin{equation}
    \boldsymbol{R}^j_i 
    = 
    \begin{pmatrix}
    \cos{\phi^j_i} & 0 & \sin{\phi^j_i} \\
    0 & 1 & 0 \\
    -\sin{\phi^j_i} & 0 & \cos{\phi^j_i} \\
    \end{pmatrix}
    \begin{pmatrix}
    1 & 0 & 0 \\
    0 & \cos{\theta^j_i} & -\sin{\theta^j_i} \\
    0 & \sin{\theta^j_i} &  \cos{\theta^j_i}, \\
    \end{pmatrix}
\end{equation}
where we define the components of the 2-dimensional condition $c^j_i = \left(\theta^j_i,\,\phi^j_i\right)$ and the dimensionality of the associated embedding $z^j_i$ becomes $N_f^j\times3$.
This is in line with the definition of spherical coordinate systems for head orientation and gaze direction estimation \cite{Zhang2018ETRA}, where the zero-representation should correspond with a ``frontal'' direction, such as the face being oriented to be directly facing the camera.
\section{Further Implementation Details}
\label{sec:Implementation}

In this section, we provide further details of the configuration and implementation of our ST-ED, and the external regression networks $F_d$ and $F_d^\prime$ for gaze estimation and head orientation. The codebase for this project can be found at \href{https://github.com/zhengyuf/ST-ED}{https://github.com/zhengyuf/ST-ED}

\subsection{Self-Transforming Encoder-Decoder (ST-ED)}

The ST-ED architecture can be flexibly configured. Here, we provide details of the backbone architecture used and the specific explicit and extraneous factors that were configured, along with their latent embedding dimensions. Lastly, we provide the used hyperparameters during training for better reproducibility.

\subsubsection{Network Architecture}
\textbf{Generator}
We use the DenseNet architecture to parameterize our encoder and decoder \cite{Huang2017Densenet}. For our decoder, we replace the convolutional layers with de-convolutions and the average-pooling layers with strided $3\times3$ de-convolutions. We configure the DenseNet with a growth rate of 32. Our input image size is $128\times128$, and we use 5 DenseNet blocks and a compression factor of 1.0. We don't use dropout or $1\times1$ convolutional layers, and use instance normalization and leaky ReLU. The feature map size at the bottleneck is $2\times2$, and we flatten the features and pass them through fully-connected layers to calculate the embeddings and pseudo-labels. Before decoding, we reshape the embeddings to have a spatial resolution of $2\times2$ to match the bottleneck's shape.

In all our experiments, we use a 0-DoF factor of size 1024, 4$\times$ 1-DoF factors of size $16\times 2$ and 4$\times$ 2-DoF factors of size $16\times 3$ for our generator. Two of the 2-DoF factors are chosen to represent the explicit factors, i.e., gaze direction and head orientation.

\textbf{Discriminator}
We use a PatchGAN discriminator as in \cite{Isola2015pix2pix}. The receptive field at the output layer is $70\times 70$, and the output size is $14 \times 14$. The architecture of the discriminator is listed in Tab. \ref{tab:our_discriminator}.
\begin{table}
\caption{Architecture of the PatchGAN discriminator used to train ST-ED \label{tab:our_discriminator}}
\begin{center}
\begin{tabular}{ |c|c| } 
 \hline
 Nr. & layers / blocks\\ 
 \hline
 0 & Conv2d(3, 64, kw=4, stride=2, pad=1, bias=True), LeakyReLU()\\ 
 \hline
 1 & Conv2d(64, 128, kw=4, stride=2, pad=1, bias=False), BatchNorm2d(128), LeakyReLU()\\ 
 \hline
 2 & Conv2d(128, 256, kw=4, stride=2, pad=1, bias=False), BatchNorm2d(256), LeakyReLU()\\ 
 \hline
 3 & Conv2d(256, 512, kw=4, stride=1, pad=1, bias=False), BatchNorm2d(512), LeakyReLU()\\ 
  \hline
 4 & Conv2d(512, 1, kw=4, stride=1, pad=1, bias=True)\\ 
 \hline
\end{tabular}
\end{center}
\end{table}
\subsubsection{Training Hyperparameters}
We use a batch size of 20, and train the network for 3 epochs (about 210k iterations). The initial learning rate is $10^{-3}$ and is decayed by $0.8$ every $0.5$ epoch. We use the Adam optimizer \cite{adamoptimizer} with a weight decay coefficient of $10^{-4}$. We use the default momentum value of $\beta_1=0.9,\, \beta_2=0.999$
\subsection{Gaze Estimation and Head Orientation Network}
We use a VGG-16 \cite{simonyan2014very} network to implement $F_d$. We select an ImageNet \cite{imagenet_cvpr09} pre-trained model and fine-tune it on the gaze and head orientation estimation tasks. The input to this network is a full-face image of size $128 \times 128$ pixels, and the output is a 4-dimensional vector representing pitch and yaw values for gaze direction and head orientation. The architecture of the network is shown in Tab. \ref{tab:external_estimator}.

We fine-tune the network for $100k$ iterations with a batch size of 64, using the Adam optimizer\cite{adamoptimizer} with momentum values $\beta_1 = 0.9,\, \beta_2 = 0.95$. The initial learning rate is $10^{-4}$ and is decayed by a factor of 0.5 after $50k$ iterations.

The external estimator for evaluation $F^\prime_d$ is trained in a similar way as $F_d$, but with a ResNet-50 \cite{ResNet} backbone which is also pre-trained on ImageNet. The architecture of the network is shown in Tab. \ref{tab:resnet_external_estimator}.
\begin{table}
\caption{Architecture of the external gaze direction and head orientation estimation network, $F_d$.\label{tab:external_estimator}}
\begin{center}
\begin{tabular}{ |c|c| } 
 \hline
 Nr. & layers / blocks\\ 
 \hline
 0 & VGG-16 convolutional layers\\ 
 \hline
 1 & FC(512, 64, w/bias), LeakyReLU()\\ 
 \hline
 2 & FC(64, 64, w/bias), LeakyReLU()\\ 
 \hline
 3 & FC(64,4, w/bias), $0.5\pi\cdot$Tanh()\\ 
 \hline
\end{tabular}
\end{center}
\end{table}

\begin{table}
\caption{Architecture of the external gaze direction and head orientation estimation network, $F^\prime_d$.\label{tab:resnet_external_estimator}
}
\begin{center}
\begin{tabular}{ |c|c| } 
 \hline
 Nr. & layers / blocks\\ 
 \hline
 0 & ResNet convolutional layers, stride of MaxPool2d = 1 \\ 
 \hline
 1 & FC(2048, 4, w/bias)\\ 

 \hline
\end{tabular}
\end{center}
\end{table}

\subsection{State-of-the-Art Baselines}

There exists no prior art in simultaneous head and gaze redirection and as such we extend and re-implement the state-of-the-art approach for gaze redirection, He\etal~\cite{He2019ICCV} and its close-cousin, StarGAN \cite{starGAN}. We perform this as best as possible by being faithful to the original objective formulations, but implement a backbone similar to our own ST-ED for fairness. This sub-section provides further details of our implementation.

\subsubsection{He\etal}
This work originally uses eye images ($64 \times 64$) from the Columbia Gaze dataset \cite{Smith2013UIST}, and performs only gaze redirection. No head orientation manipulation was shown. To apply the method to our dataset and to ensure a fair comparison, we parametrize the generator with a DenseNet-based encoder-decoder architecture, which is conceptually similar to the original down- and up-sampling generator from He\etal \cite{He2019ICCV}. We use one fewer DenseNet block compared to our ST-ED approach in both the down- and up-sampling stages of the generator, in order to match the original implementation of He\etal \cite{He2019ICCV} which has a spatially wider bottleneck than our implementation of ST-ED.
To extend the work of He\etal \cite{He2019ICCV} to perform both gaze and head redirection, we simply estimate both values with the estimation branch of the discriminator, and optimize for both.

We use a global discriminator, as in the original implementation from He\etal \cite{He2019ICCV}. The discriminator predicts a 5-dimensional vector representing the discriminator value, gaze direction and head orientation. We use a \texttt{Tanh()} function on the gaze and head direction values and then multiply by $0.5\pi$ to match the ranges of the pitch and yaw values. We follow the original implementation from He\etal \cite{He2019ICCV} and use the WGAN-GP objective \cite{wgangp}. The discriminator architecture is given in Tab.~\ref{tab:global_discriminator}.
\begin{table}:
\caption{Global discriminator network with a regression branch for gaze direction and head orientation, as used in the re-implementation of the He\etal \cite{He2019ICCV} and StarGAN \cite{starGAN} approaches.\label{tab:global_discriminator}}
\begin{center}
\begin{tabular}{ |c|c| } 
 \hline
 Nr. & layers / blocks\\ 
 \hline
 0 & Conv2d(3, 64, kw=4, stride=2, pad=1, bias=True), LeakyReLU()\\ 
 \hline
 1 & Conv2d(64, 128, kw=4, stride=2, pad=1, bias=False), LeakyReLU()\\ 
 \hline
 2 & Conv2d(128, 256, kw=4, stride=2, pad=1, bias=False), LeakyReLU()\\ 
 \hline
 3 & Conv2d(256, 512, kw=4, stride=2, pad=1, bias=False), LeakyReLU()\\ 
  \hline
   3 & Conv2d(512, 1024, kw=4, stride=2, pad=1, bias=False), LeakyReLU()\\ 
  \hline
   3 & Conv2d(1024, 2048, kw=4, stride=2, pad=1, bias=False), LeakyReLU()\\ 
  \hline
 3 & Conv2d(2048, 5, kw=2, stride=1, pad=0, bias=False)\\ 
 \hline
\end{tabular}
\end{center}
\end{table}

We increase the weight for gaze (and head orientation) estimation loss of He\etal \cite{He2019ICCV} from 5 to 1000, because gaze is harder to estimate when using full face images and the training estimation loss fails to converge when using the original weight of 5. The weights for the other loss terms are the same as in the original implementation of He\etal \cite{He2019ICCV}.
We choose the same training hyper-parameters as in our method, except that we train the network for 6 instead of 3 epochs since the WGAN-GP objective updates the generator less frequently.

\subsubsection{StarGAN}
Our implementation of StarGAN \cite{starGAN} uses the same generator and discriminator architecture as our re-implementation of He\etal \cite{He2019ICCV}. Since StarGAN does not need paired training images, we train it by redirecting the input images to random gaze and head directions, which are sampled from a 4D joint distribution of gaze and head directions computed by fitting a Gaussian kernel density to the ground truth labels from the training dataset. The weights for the cycle consistency, GAN, and gaze and head estimation losses are 400, 1 and 1000, respectively.

\subsection{Data Pre-processing}
We preprocess our image data in the same manner as done in Park\etal~\cite{Park2019ICCV}, but with changes to yield face images.
That is, we follow the pipeline defined by Zhang\etal~\cite{Zhang2018ETRA} and originally proposed by Sugano\etal~\cite{Sugano2014CVPR} but select \cite{Hu2017CVPR} and \cite{Deng2018FG} respectively for face detection and facial landmarks localization. We use the Surrey Face Model \cite{Huber2016} and the same reference 3D landmarks as in \cite{Park2019ICCV} to perform the data normalization procedure. To yield $128\times 128$ images, we configure our virtual camera to have a focal length of $500mm$ and distance-to-subject of $600mm$.

In more simple terms, we use the code\footnote{\url{https://github.com/swook/faze_preprocess}} provided by Park\etal~\cite{Park2019ICCV} and tweak the parameters in the ``normalized\_camera'' variable.

\begin{figure}[ht]
\centering
\begin{subfigure}{.15\linewidth}
\includegraphics[width=\linewidth]{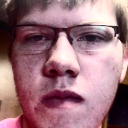}
\end{subfigure}
\begin{subfigure}{.15\linewidth}
\includegraphics[width=\linewidth]{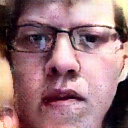}
\end{subfigure}
\begin{subfigure}{.15\linewidth}
\includegraphics[width=\linewidth]{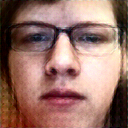}
\end{subfigure}
\begin{subfigure}{.15\linewidth}
\includegraphics[width=\linewidth]{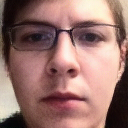}
\end{subfigure}
\begin{subfigure}{.15\linewidth}
\includegraphics[width=\linewidth]{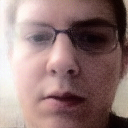}
\end{subfigure}
\begin{subfigure}{.15\linewidth}
\includegraphics[width=\linewidth]{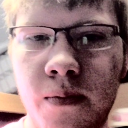}
\end{subfigure}
\begin{subfigure}{.15\linewidth}
\includegraphics[width=\linewidth]{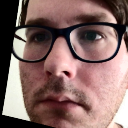}
\end{subfigure}
\begin{subfigure}{.15\linewidth}
\includegraphics[width=\linewidth]{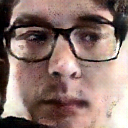}
\end{subfigure}
\begin{subfigure}{.15\linewidth}
\includegraphics[width=\linewidth]{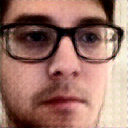}
\end{subfigure}
\begin{subfigure}{.15\linewidth}
\includegraphics[width=\linewidth]{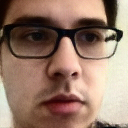}
\end{subfigure}
\begin{subfigure}{.15\linewidth}
\includegraphics[width=\linewidth]{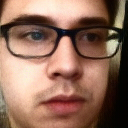}
\end{subfigure}
\begin{subfigure}{.15\linewidth}
\includegraphics[width=\linewidth]{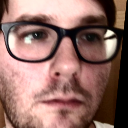}
\end{subfigure}
\\
\begin{subfigure}{.15\linewidth}
\includegraphics[width=\linewidth]{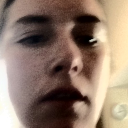}
\end{subfigure}
\begin{subfigure}{.15\linewidth}
\includegraphics[width=\linewidth]{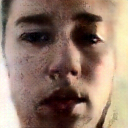}
\end{subfigure}
\begin{subfigure}{.15\linewidth}
\includegraphics[width=\linewidth]{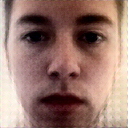}
\end{subfigure}
\begin{subfigure}{.15\linewidth}
\includegraphics[width=\linewidth]{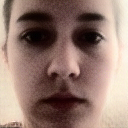}
\end{subfigure}
\begin{subfigure}{.15\linewidth}
\includegraphics[width=\linewidth]{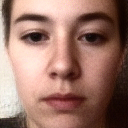}
\end{subfigure}
\begin{subfigure}{.15\linewidth}
\includegraphics[width=\linewidth]{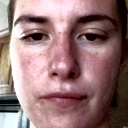}
\end{subfigure}
\\
\begin{subfigure}{.15\linewidth}
\includegraphics[width=\linewidth]{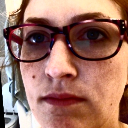}
\end{subfigure}
\begin{subfigure}{.15\linewidth}
\includegraphics[width=\linewidth]{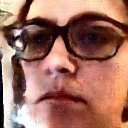}
\end{subfigure}
\begin{subfigure}{.15\linewidth}
\includegraphics[width=\linewidth]{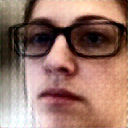}
\end{subfigure}
\begin{subfigure}{.15\linewidth}
\includegraphics[width=\linewidth]{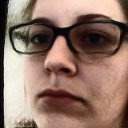}
\end{subfigure}
\begin{subfigure}{.15\linewidth}
\includegraphics[width=\linewidth]{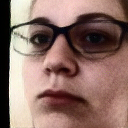}
\end{subfigure}
\begin{subfigure}{.15\linewidth}
\includegraphics[width=\linewidth]{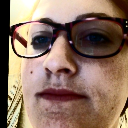}
\end{subfigure}
\\
\begin{subfigure}{.15\linewidth}
\includegraphics[width=\linewidth]{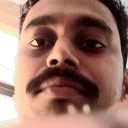}
\end{subfigure}
\begin{subfigure}{.15\linewidth}
\includegraphics[width=\linewidth]{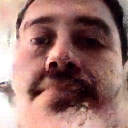}
\end{subfigure}
\begin{subfigure}{.15\linewidth}
\includegraphics[width=\linewidth]{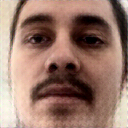}
\end{subfigure}
\begin{subfigure}{.15\linewidth}
\includegraphics[width=\linewidth]{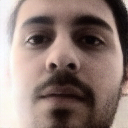}
\end{subfigure}
\begin{subfigure}{.15\linewidth}
\includegraphics[width=\linewidth]{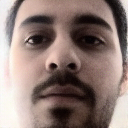}
\end{subfigure}
\begin{subfigure}{.15\linewidth}
\includegraphics[width=\linewidth]{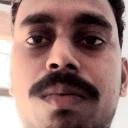}
\end{subfigure}
\\
\begin{subfigure}{.15\linewidth}
\includegraphics[width=\linewidth]{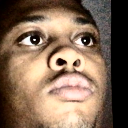}
\end{subfigure}
\begin{subfigure}{.15\linewidth}
\includegraphics[width=\linewidth]{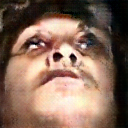}
\end{subfigure}
\begin{subfigure}{.15\linewidth}
\includegraphics[width=\linewidth]{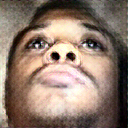}
\end{subfigure}
\begin{subfigure}{.15\linewidth}
\includegraphics[width=\linewidth]{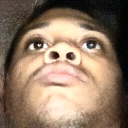}
\end{subfigure}
\begin{subfigure}{.15\linewidth}
\includegraphics[width=\linewidth]{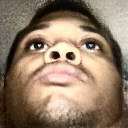}
\end{subfigure}
\begin{subfigure}{.15\linewidth}
\includegraphics[width=\linewidth]{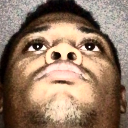}
\end{subfigure}
\\
\begin{subfigure}{.15\linewidth}
\includegraphics[width=\linewidth]{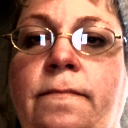}
\end{subfigure}
\begin{subfigure}{.15\linewidth}
\includegraphics[width=\linewidth]{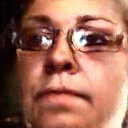}
\end{subfigure}
\begin{subfigure}{.15\linewidth}
\includegraphics[width=\linewidth]{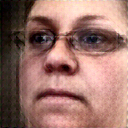}
\end{subfigure}
\begin{subfigure}{.15\linewidth}
\includegraphics[width=\linewidth]{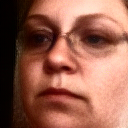}
\end{subfigure}
\begin{subfigure}{.15\linewidth}
\includegraphics[width=\linewidth]{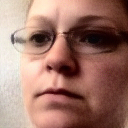}
\end{subfigure}
\begin{subfigure}{.15\linewidth}
\includegraphics[width=\linewidth]{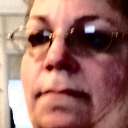}
\end{subfigure}
\\
\begin{subfigure}{.15\linewidth}
\includegraphics[width=\linewidth]{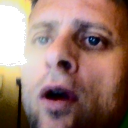}
\end{subfigure}
\begin{subfigure}{.15\linewidth}
\includegraphics[width=\linewidth]{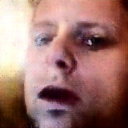}
\end{subfigure}
\begin{subfigure}{.15\linewidth}
\includegraphics[width=\linewidth]{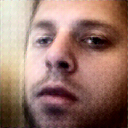}
\end{subfigure}
\begin{subfigure}{.15\linewidth}
\includegraphics[width=\linewidth]{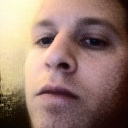}
\end{subfigure}
\begin{subfigure}{.15\linewidth}
\includegraphics[width=\linewidth]{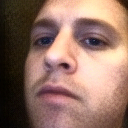}
\end{subfigure}
\begin{subfigure}{.15\linewidth}
\includegraphics[width=\linewidth]{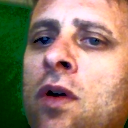}
\end{subfigure}
\\
\subcaptionbox{Input image}[.15\linewidth]{\includegraphics[width=\linewidth]{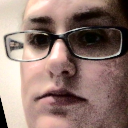}}
\subcaptionbox{StarGAN \cite{starGAN}}[.15\linewidth]{\includegraphics[width=\linewidth]{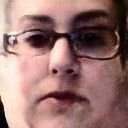}}
\subcaptionbox{He et al. \cite{He2019ICCV}}[.15\linewidth]{\includegraphics[width=\linewidth]{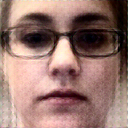}}
\subcaptionbox{Ours ($g+h$)
}[.15\linewidth]{\includegraphics[width=\linewidth]{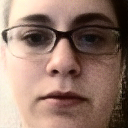}}
\subcaptionbox{Ours (all)
}[.15\linewidth]{\includegraphics[width=\linewidth]{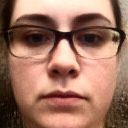}}
\subcaptionbox{Target Image}[.15\linewidth]{\includegraphics[width=\linewidth]{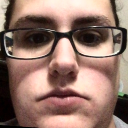}}
\caption{continued on the next page...}
\end{figure}
\begin{figure}\ContinuedFloat
\centering
\begin{subfigure}{.15\linewidth}
\includegraphics[width=\linewidth]{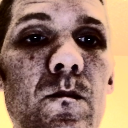}
\end{subfigure}
\begin{subfigure}{.15\linewidth}
\includegraphics[width=\linewidth]{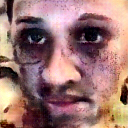}
\end{subfigure}
\begin{subfigure}{.15\linewidth}
\includegraphics[width=\linewidth]{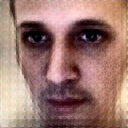}
\end{subfigure}
\begin{subfigure}{.15\linewidth}
\includegraphics[width=\linewidth]{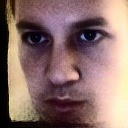}
\end{subfigure}
\begin{subfigure}{.15\linewidth}
\includegraphics[width=\linewidth]{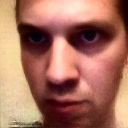}
\end{subfigure}
\begin{subfigure}{.15\linewidth}
\includegraphics[width=\linewidth]{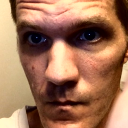}
\end{subfigure}
\\
\begin{subfigure}{.15\linewidth}
\includegraphics[width=\linewidth]{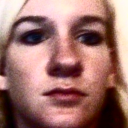}
\end{subfigure}
\begin{subfigure}{.15\linewidth}
\includegraphics[width=\linewidth]{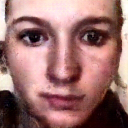}
\end{subfigure}
\begin{subfigure}{.15\linewidth}
\includegraphics[width=\linewidth]{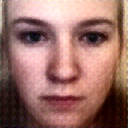}
\end{subfigure}
\begin{subfigure}{.15\linewidth}
\includegraphics[width=\linewidth]{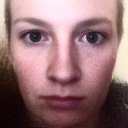}
\end{subfigure}
\begin{subfigure}{.15\linewidth}
\includegraphics[width=\linewidth]{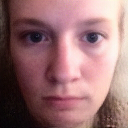}
\end{subfigure}
\begin{subfigure}{.15\linewidth}
\includegraphics[width=\linewidth]{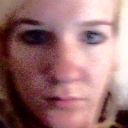}
\end{subfigure}
\\
\begin{subfigure}{.15\linewidth}
\includegraphics[width=\linewidth]{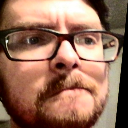}
\end{subfigure}
\begin{subfigure}{.15\linewidth}
\includegraphics[width=\linewidth]{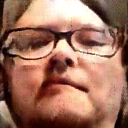}
\end{subfigure}
\begin{subfigure}{.15\linewidth}
\includegraphics[width=\linewidth]{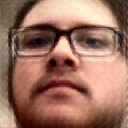}
\end{subfigure}
\begin{subfigure}{.15\linewidth}
\includegraphics[width=\linewidth]{figures/qualitative_sota/953/ours.png}
\end{subfigure}
\begin{subfigure}{.15\linewidth}
\includegraphics[width=\linewidth]{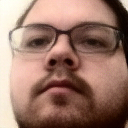}
\end{subfigure}
\begin{subfigure}{.15\linewidth}
\includegraphics[width=\linewidth]{figures/qualitative_sota/953/target_image.png}
\end{subfigure}
\\
\begin{subfigure}{.15\linewidth}
\includegraphics[width=\linewidth]{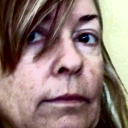}
\end{subfigure}
\begin{subfigure}{.15\linewidth}
\includegraphics[width=\linewidth]{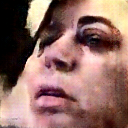}
\end{subfigure}
\begin{subfigure}{.15\linewidth}
\includegraphics[width=\linewidth]{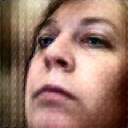}
\end{subfigure}
\begin{subfigure}{.15\linewidth}
\includegraphics[width=\linewidth]{figures/qualitative_sota/1575/ours.png}
\end{subfigure}
\begin{subfigure}{.15\linewidth}
\includegraphics[width=\linewidth]{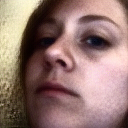}
\end{subfigure}
\begin{subfigure}{.15\linewidth}
\includegraphics[width=\linewidth]{figures/qualitative_sota/1575/target_image.png}
\end{subfigure}
\\
\begin{subfigure}{.15\linewidth}
\includegraphics[width=\linewidth]{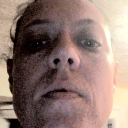}
\end{subfigure}
\begin{subfigure}{.15\linewidth}
\includegraphics[width=\linewidth]{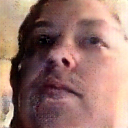}
\end{subfigure}
\begin{subfigure}{.15\linewidth}
\includegraphics[width=\linewidth]{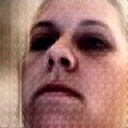}
\end{subfigure}
\begin{subfigure}{.15\linewidth}
\includegraphics[width=\linewidth]{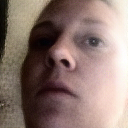}
\end{subfigure}
\begin{subfigure}{.15\linewidth}
\includegraphics[width=\linewidth]{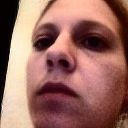}
\end{subfigure}
\begin{subfigure}{.15\linewidth}
\includegraphics[width=\linewidth]{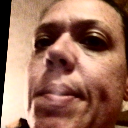}
\end{subfigure}
\\
\begin{subfigure}{.15\linewidth}
\includegraphics[width=\linewidth]{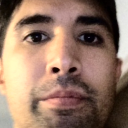}
\end{subfigure}
\begin{subfigure}{.15\linewidth}
\includegraphics[width=\linewidth]{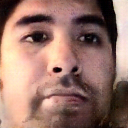}
\end{subfigure}
\begin{subfigure}{.15\linewidth}
\includegraphics[width=\linewidth]{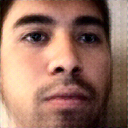}
\end{subfigure}
\begin{subfigure}{.15\linewidth}
\includegraphics[width=\linewidth]{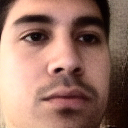}
\end{subfigure}
\begin{subfigure}{.15\linewidth}
\includegraphics[width=\linewidth]{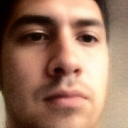}
\end{subfigure}
\begin{subfigure}{.15\linewidth}
\includegraphics[width=\linewidth]{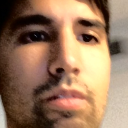}
\end{subfigure}
\\
\begin{subfigure}{.15\linewidth}
\includegraphics[width=\linewidth]{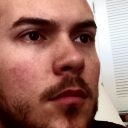}
\end{subfigure}
\begin{subfigure}{.15\linewidth}
\includegraphics[width=\linewidth]{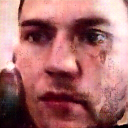}
\end{subfigure}
\begin{subfigure}{.15\linewidth}
\includegraphics[width=\linewidth]{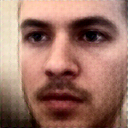}
\end{subfigure}
\begin{subfigure}{.15\linewidth}
\includegraphics[width=\linewidth]{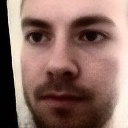}
\end{subfigure}
\begin{subfigure}{.15\linewidth}
\includegraphics[width=\linewidth]{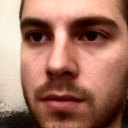}
\end{subfigure}
\begin{subfigure}{.15\linewidth}
\includegraphics[width=\linewidth]{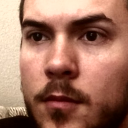}
\end{subfigure}
\\
\begin{subfigure}{.15\linewidth}
\includegraphics[width=\linewidth]{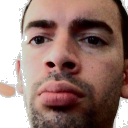}
\end{subfigure}
\begin{subfigure}{.15\linewidth}
\includegraphics[width=\linewidth]{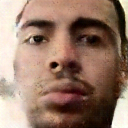}
\end{subfigure}
\begin{subfigure}{.15\linewidth}
\includegraphics[width=\linewidth]{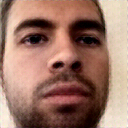}
\end{subfigure}
\begin{subfigure}{.15\linewidth}
\includegraphics[width=\linewidth]{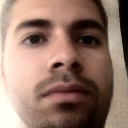}
\end{subfigure}
\begin{subfigure}{.15\linewidth}
\includegraphics[width=\linewidth]{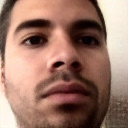}
\end{subfigure}
\begin{subfigure}{.15\linewidth}
\includegraphics[width=\linewidth]{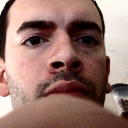}
\end{subfigure}
\setcounter{subfigure}{0}
\subcaptionbox{Input image}[.15\linewidth]{\includegraphics[width=\linewidth]{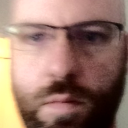}}
\subcaptionbox{StarGAN \cite{starGAN}}[.15\linewidth]{\includegraphics[width=\linewidth]{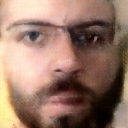}}
\subcaptionbox{He et al. \cite{He2019ICCV}}[.15\linewidth]{\includegraphics[width=\linewidth]{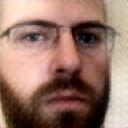}}
\subcaptionbox{Ours ($g+h$)
}[.15\linewidth]{\includegraphics[width=\linewidth]{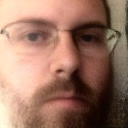}}
\subcaptionbox{Ours (all)
}[.15\linewidth]{\includegraphics[width=\linewidth]{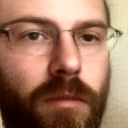}}
\subcaptionbox{Target Image}[.15\linewidth]{\includegraphics[width=\linewidth]{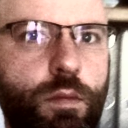}}

\caption{
    \textbf{Qualitative Results.}
    Example redirection results on the test subset of GazeCapture \cite{Krafka2016CVPR}.
    Our method produces more detailed and photo-realistic images compared to the baseline methods of He\etal \cite{He2019ICCV} and StarGAN \cite{starGAN}. Note that our method can generate photo-realistic images even in cases of large head pose changes, eyeglasses, and blurry inputs. By aligning to all predicted pseudo-labels of a target ground-truth image, our approach can also better approximate the target.
}
\label{fig:additional qualitative results}
\end{figure}
\clearpage
\small
\bibliographystyle{unsrt}
\bibliography{bibliography}

\begin{thebibliography}{10}

\bibitem{Oertel2015ICMI}
Catharine Oertel, Kenneth~A Funes~Mora, Joakim Gustafson, and Jean-Marc Odobez.
\newblock Deciphering the silent participant: On the use of audio-visual cues
  for the classification of listener categories in group discussions.
\newblock In {\em Proceedings of the 2015 ACM on International Conference on
  Multimodal Interaction}, pages 107--114, 2015.

\bibitem{Fan2019ICCV}
Lifeng Fan, Wenguan Wang, Siyuan Huang, Xinyu Tang, and Song-Chun Zhu.
\newblock Understanding human gaze communication by spatio-temporal graph
  reasoning.
\newblock In {\em ICCV}, pages 5724--5733, 2019.

\bibitem{buswell1935people}
Guy~Thomas Buswell.
\newblock How people look at pictures: a study of the psychology and perception
  in art.
\newblock 1935.

\bibitem{rothkopf2007task}
Constantin~A Rothkopf, Dana~H Ballard, and Mary~M Hayhoe.
\newblock Task and context determine where you look.
\newblock {\em Journal of vision}, 7(14):16--16, 2007.

\bibitem{Fridman2018CHI}
Lex Fridman, Bryan Reimer, Bruce Mehler, and William~T. Freeman.
\newblock Cognitive load estimation in the wild.
\newblock In {\em ACM CHI}, 2018.

\bibitem{Huang2016MM}
Michael~Xuelin Huang, Jiajia Li, Grace Ngai, and Hong~Va Leong.
\newblock Stressclick: Sensing stress from gaze-click patterns.
\newblock In {\em ACM MM}, 2016.

\bibitem{Feit2017CHI}
Anna~Maria Feit, Shane Williams, Arturo Toledo, Ann Paradiso, Harish Kulkarni,
  Shaun~K. Kane, and Meredith~Ringel Morris.
\newblock Toward everyday gaze input: Accuracy and precision of eye tracking
  and implications for design.
\newblock In {\em ACM CHI}, pages 1118--1130, 2017.

\bibitem{Smith2013UIST}
B.A. Smith, Q.~Yin, S.K. Feiner, and S.K. Nayar.
\newblock {G}aze {L}ocking: {P}assive {E}ye {C}ontact {D}etection for
  {H}uman-{O}bject {I}nteraction.
\newblock In {\em ACM UIST}, pages 271--280, Oct 2013.

\bibitem{Zhang2017UIST}
Xucong Zhang, Yusuke Sugano, and Andreas Bulling.
\newblock Everyday eye contact detection using unsupervised gaze target
  discovery.
\newblock In {\em ACM UIST}, pages 193--203, 2017.

\bibitem{Patney2016SIGGRAPH}
Anjul Patney, Joohwan Kim, Marco Salvi, Anton Kaplanyan, Chris Wyman, Nir
  Benty, Aaron Lefohn, and David Luebke.
\newblock Perceptually-based foveated virtual reality.
\newblock In {\em SIGGRAPH}, 2016.

\bibitem{Wood2015ICCV}
Erroll Wood, Tadas Baltruaitis, Xucong Zhang, Yusuke Sugano, Peter Robinson,
  and Andreas Bulling.
\newblock Rendering of eyes for eye-shape registration and gaze estimation.
\newblock In {\em ICCV}, pages 3756--3764, 2015.

\bibitem{Wood2016ETRA}
Erroll Wood, Tadas Baltru\v{s}aitis, Louis-Philippe Morency, Peter Robinson,
  and Andreas Bulling.
\newblock Learning an appearance-based gaze estimator from one million
  synthesised images.
\newblock In {\em ACM ETRA}, page 131–138, New York, NY, USA, 2016.
  Association for Computing Machinery.

\bibitem{Shrivastava2017CVPR}
Ashish Shrivastava, Tomas Pfister, Oncel Tuzel, Joshua Susskind, Wenda Wang,
  and Russell Webb.
\newblock Learning from simulated and unsupervised images through adversarial
  training.
\newblock In {\em CVPR}, July 2017.

\bibitem{Lee2018ICLR}
Kangwook Lee, Hoon Kim, and Changho Suh.
\newblock Simulated+unsupervised learning with adaptive data generation and
  bidirectional mappings.
\newblock In {\em ICLR}, 2018.

\bibitem{Ganin2016ECCV}
Yaroslav Ganin, Daniil Kononenko, Diana Sungatullina, and Victor Lempitsky.
\newblock Deepwarp: Photorealistic image resynthesis for gaze manipulation.
\newblock In {\em ECCV}, pages 311--326. Springer, 2016.

\bibitem{Yu2019CVPR}
Yu~Yu, Gang Liu, and Jean-Marc Odobez.
\newblock Improving few-shot user-specific gaze adaptation via gaze redirection
  synthesis.
\newblock In {\em CVPR}, June 2019.

\bibitem{He2019ICCV}
Zhe He, Adrian Spurr, Xucong Zhang, and Otmar Hilliges.
\newblock Photo-realistic monocular gaze redirection using generative
  adversarial networks.
\newblock In {\em ICCV}. {IEEE}, 2019.

\bibitem{Krafka2016CVPR}
Kyle Krafka, Aditya Khosla, Petr Kellnhofer, Harini Kannan, Suchendra
  Bhandarkar, Wojciech Matusik, and Antonio Torralba.
\newblock {Eye Tracking for Everyone}.
\newblock In {\em CVPR}, 2016.

\bibitem{Zhang2015CVPR}
Xucong Zhang, Yusuke Sugano, Mario Fritz, and Andreas Bulling.
\newblock Appearance-based gaze estimation in the wild.
\newblock In {\em CVPR}, 2015.

\bibitem{Park2019ICCV}
Seonwook Park, Shalini~De Mello, Pavlo Molchanov, Umar Iqbal, Otmar Hilliges,
  and Jan Kautz.
\newblock Few-shot adaptive gaze estimation.
\newblock In {\em ICCV}, 2019.

\bibitem{starGAN}
Yunjey Choi, Minje Choi, Munyoung Kim, Jung-Woo Ha, Sunghun Kim, and Jaegul
  Choo.
\newblock Stargan: Unified generative adversarial networks for multi-domain
  image-to-image translation.
\newblock In {\em CVPR}, pages 8789--8797, 2018.

\bibitem{he2019attgan}
Zhenliang He, Wangmeng Zuo, Meina Kan, Shiguang Shan, and Xilin Chen.
\newblock Attgan: Facial attribute editing by only changing what you want.
\newblock {\em TIP}, 28(11):5464--5478, 2019.

\bibitem{pumarola2018ganimation}
Albert Pumarola, Antonio Agudo, Aleix~M Martinez, Alberto Sanfeliu, and
  Francesc Moreno-Noguer.
\newblock Ganimation: Anatomically-aware facial animation from a single image.
\newblock In {\em ECCV}, pages 818--833, 2018.

\bibitem{wu2019relgan}
Po-Wei Wu, Yu-Jing Lin, Che-Han Chang, Edward~Y Chang, and Shih-Wei Liao.
\newblock Relgan: Multi-domain image-to-image translation via relative
  attributes.
\newblock In {\em ICCV}, pages 5914--5922, 2019.

\bibitem{wood2016learning}
Erroll Wood, Tadas Baltru{\v{s}}aitis, Louis-Philippe Morency, Peter Robinson,
  and Andreas Bulling.
\newblock Learning an appearance-based gaze estimator from one million
  synthesised images.
\newblock In {\em ACM ETRA}, pages 131--138, 2016.

\bibitem{johnson2016perceptual}
Justin Johnson, Alexandre Alahi, and Li~Fei-Fei.
\newblock Perceptual losses for real-time style transfer and super-resolution.
\newblock In {\em ECCV}, pages 694--711. Springer, 2016.

\bibitem{cycleGAN}
Jun-Yan Zhu, Taesung Park, Phillip Isola, and Alexei~A Efros.
\newblock Unpaired image-to-image translation using cycle-consistent
  adversarial networks.
\newblock In {\em ICCV}, pages 2223--2232, 2017.

\bibitem{wood2018gazedirector}
Erroll Wood, Tadas Baltru{\v{s}}aitis, Louis-Philippe Morency, Peter Robinson,
  and Andreas Bulling.
\newblock Gazedirector: Fully articulated eye gaze redirection in video.
\newblock In {\em Computer Graphics Forum}, volume~37, pages 217--225. Wiley
  Online Library, 2018.

\bibitem{Zhang2019TPAMI}
Xucong Zhang, Yusuke Sugano, Mario Fritz, and Andreas Bulling.
\newblock Mpiigaze: Real-world dataset and deep appearance-based gaze
  estimation.
\newblock {\em TPAMI}, 2019.

\bibitem{Wang2018CVPR}
Kang Wang, Rui Zhao, and Qiang Ji.
\newblock A hierarchical generative model for eye image synthesis and eye gaze
  estimation.
\newblock In {\em CVPR}, 2018.

\bibitem{Park2018ETRA}
Seonwook Park, Xucong Zhang, Andreas Bulling, and Otmar Hilliges.
\newblock Learning to find eye region landmarks for remote gaze estimation in
  unconstrained settings.
\newblock In {\em ACM ETRA}, 2018.

\bibitem{Huang2018ECCV}
Yifei Huang, Minjie Cai, Zhenqiang Li, and Yoichi Sato.
\newblock Predicting gaze in egocentric video by learning task-dependent
  attention transition.
\newblock In {\em ECCV}, 2018.

\bibitem{park2019semantic}
Taesung Park, Ming-Yu Liu, Ting-Chun Wang, and Jun-Yan Zhu.
\newblock Semantic image synthesis with spatially-adaptive normalization.
\newblock In {\em CVPR}, pages 2337--2346, 2019.

\bibitem{lee2018diverse}
Hsin-Ying Lee, Hung-Yu Tseng, Jia-Bin Huang, Maneesh Singh, and Ming-Hsuan
  Yang.
\newblock Diverse image-to-image translation via disentangled representations.
\newblock In {\em ECCV}, pages 35--51, 2018.

\bibitem{mejjati2018unsupervised}
Youssef~Alami Mejjati, Christian Richardt, James Tompkin, Darren Cosker, and
  Kwang~In Kim.
\newblock Unsupervised attention-guided image-to-image translation.
\newblock In {\em NeurIPS}, pages 3693--3703, 2018.

\bibitem{configECCV2020}
Marek Kowalski, Stephan~J. Garbin, Virginia Estellers, Tadas Baltrušaitis,
  Matthew Johnson, and Jamie Shotton.
\newblock Config: Controllable neural face image generation.
\newblock In {\em European Conference on Computer Vision (ECCV)}, 2020.

\bibitem{mirza2014conditional}
Mehdi Mirza and Simon Osindero.
\newblock Conditional generative adversarial nets.
\newblock {\em arXiv preprint arXiv:1411.1784}, 2014.

\bibitem{Hinton2011ICANN}
Geoffrey~E Hinton, Alex Krizhevsky, and Sida~D Wang.
\newblock Transforming auto-encoders.
\newblock In {\em ICANN}, 2011.

\bibitem{Chen2019ICCV}
Xu~Chen, Jie Song, and Otmar Hilliges.
\newblock Monocular neural image based rendering with continuous view control.
\newblock In {\em ICCV}, October 2019.

\bibitem{mustikovelaCVPR20}
Siva~Karthik Mustikovela, Varun Jampani, Shalini De~Mello, Sifei Liu, Umar
  Iqbal, Carsten Rother, and Jan Kautz.
\newblock Self-supervised viewpoint learning from image collections.
\newblock In {\em CVPR}, 2020.

\bibitem{Nguyen2019ICCV}
Thu Nguyen-Phuoc, Chuan Li, Lucas Theis, Christian Richardt, and Yong-Liang
  Yang.
\newblock Hologan: Unsupervised learning of 3d representations from natural
  images.
\newblock In {\em ICCV}, pages 7588--7597, 2019.

\bibitem{Worrall2017ICCV}
Daniel~E Worrall, Stephan~J Garbin, Daniyar Turmukhambetov, and Gabriel~J
  Brostow.
\newblock Interpretable transformations with encoder-decoder networks.
\newblock In {\em ICCV}, 2017.

\bibitem{Rhodin2018ECCV}
Helge Rhodin, Mathieu Salzmann, and Pascal Fua.
\newblock Unsupervised geometry-aware representation for 3d human pose
  estimation.
\newblock In {\em ECCV}, 2018.

\bibitem{Dosovitskiy2016}
Alexey Dosovitskiy and Thomas Brox.
\newblock Generating images with perceptual similarity metrics based on deep
  networks.
\newblock In D.~D. Lee, M.~Sugiyama, U.~V. Luxburg, I.~Guyon, and R.~Garnett,
  editors, {\em NeurIPS}, pages 658--666. 2016.

\bibitem{goodfellow2014generative}
Ian Goodfellow, Jean Pouget-Abadie, Mehdi Mirza, Bing Xu, David Warde-Farley,
  Sherjil Ozair, Aaron Courville, and Yoshua Bengio.
\newblock Generative adversarial nets.
\newblock In {\em NeurIPS}, pages 2672--2680, 2014.

\bibitem{Isola2015pix2pix}
P.~{Isola}, J.~{Zhu}, T.~{Zhou}, and A.~A. {Efros}.
\newblock Image-to-image translation with conditional adversarial networks.
\newblock In {\em CVPR}, 2017.

\bibitem{simonyan2014very}
Karen Simonyan and Andrew Zisserman.
\newblock Very deep convolutional networks for large-scale image recognition.
\newblock In {\em ICLR}, 2014.

\bibitem{ResNet}
Kaiming He, Xiangyu Zhang, Shaoqing Ren, and Jian Sun.
\newblock Deep residual learning for image recognition.
\newblock {\em CoRR}, abs/1512.03385, 2015.

\bibitem{Sugano2014CVPR}
Yusuke Sugano, Yasuyuki Matsushita, and Yoichi Sato.
\newblock {Learning-by-Synthesis for Appearance-based 3D Gaze Estimation}.
\newblock In {\em CVPR}, 2014.

\bibitem{Zhang2018ETRA}
Xucong Zhang, Yusuke Sugano, and Andreas Bulling.
\newblock Revisiting data normalization for appearance-based gaze estimation.
\newblock In {\em ETRA}, 2018.

\bibitem{Zhang2017CVPRW}
Xucong Zhang, Yusuke Sugano, Mario Fritz, and Andreas Bulling.
\newblock It's written all over your face: Full-face appearance-based gaze
  estimation.
\newblock In {\em CVPRW}, 2017.

\bibitem{FunesMora2014ETRA}
Kenneth~Alberto Funes~Mora, Florent Monay, and Jean-Marc Odobez.
\newblock Eyediap: A database for the development and evaluation of gaze
  estimation algorithms from rgb and rgb-d cameras.
\newblock In {\em ACM ETRA}. ACM, March 2014.

\bibitem{higgins2017beta}
Irina Higgins, Loic Matthey, Arka Pal, Christopher Burgess, Xavier Glorot,
  Matthew Botvinick, Shakir Mohamed, and Alexander Lerchner.
\newblock beta-vae: Learning basic visual concepts with a constrained
  variational framework.
\newblock In {\em ICLR}, 2017.

\bibitem{Zhang2018lpips}
Richard Zhang, Phillip Isola, Alexei~A Efros, Eli Shechtman, and Oliver Wang.
\newblock The unreasonable effectiveness of deep features as a perceptual
  metric.
\newblock In {\em CVPR}, pages 586--595, 2018.

\bibitem{Huang2017Densenet}
G.~{Huang}, Z.~{Liu}, L.~{Van Der Maaten}, and K.~Q. {Weinberger}.
\newblock Densely connected convolutional networks.
\newblock In {\em CVPR}, pages 2261--2269, 2017.

\bibitem{adamoptimizer}
Diederik Kingma and Jimmy Ba.
\newblock Adam: A method for stochastic optimization.
\newblock {\em ICLR}, 12 2014.

\bibitem{imagenet_cvpr09}
J.~Deng, W.~Dong, R.~Socher, L.-J. Li, K.~Li, and L.~Fei-Fei.
\newblock {ImageNet: A Large-Scale Hierarchical Image Database}.
\newblock In {\em CVPR}, 2009.

\bibitem{wgangp}
Ishaan Gulrajani, Faruk Ahmed, Martin Arjovsky, Vincent Dumoulin, and Aaron~C
  Courville.
\newblock Improved training of wasserstein gans.
\newblock In I.~Guyon, U.~V. Luxburg, S.~Bengio, H.~Wallach, R.~Fergus,
  S.~Vishwanathan, and R.~Garnett, editors, {\em NeurIPS}, 2017.

\bibitem{Hu2017CVPR}
Peiyun Hu and Deva Ramanan.
\newblock Finding tiny faces.
\newblock In {\em CVPR}, 2017.

\bibitem{Deng2018FG}
J.~{Deng}, Y.~{Zhou}, S.~{Cheng}, and S.~{Zaferiou}.
\newblock Cascade multi-view hourglass model for robust 3d face alignment.
\newblock In {\em FG}, 2018.

\bibitem{Huber2016}
Patrik Huber, Guosheng Hu, Rafael Tena, Pouria Mortazavian, P~Koppen, William~J
  Christmas, Matthias Ratsch, and Josef Kittler.
\newblock A multiresolution 3d morphable face model and fitting framework.
\newblock In {\em VISIGRAPP}, 2016.

\end{thebibliography}

\end{document}